\documentclass[twocolumn]{IEEEtran}

\usepackage[utf8]{inputenc} 
\usepackage[T1]{fontenc}    
\usepackage{url}            
\usepackage{booktabs}       
\usepackage{microtype} 
\usepackage{paralist} 
\usepackage{cellspace}
\usepackage[ruled,vlined]{algorithm2e}
\usepackage{bm}
\usepackage{amsfonts}  
\usepackage{amssymb}   
\usepackage{amsmath}
\usepackage{mathrsfs}
\usepackage[pdftex]{graphicx}
\usepackage[subrefformat=parens,labelformat=parens]{subfig}

\usepackage{tabularx}
\usepackage[flushleft]{threeparttable} 
\usepackage{makecell}
\usepackage{multirow}

\usepackage{nicefrac}       
\usepackage{hyperref}
\hypersetup{
    colorlinks=true,
    linkcolor=blue,
    filecolor=magenta,      
    urlcolor=cyan,
}

\usepackage{cleveref}  
\crefformat{table}{Table~#2#1#3}
\crefformat{section}{Section~#2#1#3}
\crefformat{figure}{Fig.~#2#1#3}
\crefformat{equation}{Eq.~(#2#1#3)}
\crefformat{algorithm}{Algorithm~#2#1#3}
\crefformat{appendix}{Appendix #2#1#3}
\usepackage[numbers,sort&compress]{natbib}

\usepackage{xspace}

\newcommand{\etal}{\emph{et al.}\xspace}

\begin{document}

\setlength{\abovedisplayskip}{.5\baselineskip}
\setlength{\belowdisplayskip}{.5\baselineskip}

\title{Attentional Local Contrast Networks for Infrared Small Target Detection}

\author{
Yimian~Dai, Yiquan~Wu, Fei~Zhou, Kobus~Barnard
    \thanks{Y. Dai, Y. Wu, and F. Zhou are with the College of Electronic and Information Engineering, Nanjing University of Aeronautics and Astronautics, Nanjing 211106, China (e-mail: \{yimian.dai, nuaaimage\}@gmail.com; hellozf1990@163.com).}
    
    \thanks{K. Barnard is with the Department of Computer Science, University of Arizona, Tucson 85721, US (e-mail: kobus@cs.arizona.edu).}
}

\maketitle

\begin{abstract}
To mitigate the issue of minimal intrinsic features for pure data-driven methods, in this paper, we propose a novel model-driven deep network for infrared small target detection, which combines discriminative networks and conventional model-driven methods to make use of both labeled data and the domain knowledge.
By designing a feature map cyclic shift scheme, we modularize a conventional local contrast measure method as a depth-wise parameterless nonlinear feature refinement layer in an end-to-end network, which encodes relatively long-range contextual interactions with clear physical interpretability.
To highlight and preserve the small target features, we also exploit a bottom-up attentional modulation integrating the smaller scale subtle details of low-level features into high-level features of deeper layers.
We conduct detailed ablation studies with varying network depths to empirically verify the effectiveness and efficiency of the design of each component in our network architecture.
We also compare the performance of our network against other model-driven methods and deep networks on the open SIRST dataset as well.
The results suggest that our network yields a performance boost over its competitors.
Our code, trained models, and results are available online\footnote{\url{https://github.com/YimianDai/open-alcnet}}.

\end{abstract}  


\section{Introduction}

\IEEEPARstart{I}{nfrared} small target detection plays an important role in applications like early-warning systems and maritime surveillance systems because of the ability of infrared imaging that can offer a clear image without illumination or penetrate obstructions like fog, smoke, and other atmospheric conditions~\cite{CVPRW14TIVBenchmark}.
Due to the long imaging distance, the infrared target generally ends up only occupying a few pixels in the image, lacking texture or shape characteristics~\cite{ICNNSP03SPIE}.
Facing the lack of intrinsic features, the traditional methods utilize the spatio-temporal continuity of the target on the image sequence for detection by assuming a static background or consistent targets in adjacent frames~\cite{TAES88MatchedFiltering}.
Recently, the need for early warning has made researchers pay more and more attention to the \emph{single-frame detection} task, particularly with the advance in hypersonic aircraft in which the fast-changing backgrounds and inconsistent target motion traces caused by the rapid relative movement between sensor platforms and targets can make the performance of sequential detection methods degrade significantly~\cite{TIP13IPI}.
Therefore, detecting small targets in a single image is of great importance for such applications.

Conventional \emph{model-driven} approaches model the infrared small targets as an outlier popping out from the slowly transitional background where nearby pixels are highly correlated~\cite{TASS87MaxMedian}.
Detecting infrared small targets is thus a form of blob detection, which is a problem with a long history in the image processing literature~\cite{IJCV04BlobDetection}. 
However, in real-world scenarios, the problem is more complex: there are more distractors that also stand out as outliers in the background~\cite{JSTARS18GVF}. 
Hence, model-driven methods must make strong prior assumptions about the small target, e.g., the most sparse~\cite{TIP13IPI} or salient~\cite{TGRS13LCM}, about the background, e.g., smooth~\cite{TGRS16WLDM} or non-local correlated~\cite{IPT17NIPPS}, or both~\cite{JSTARS17RIPT}.
Traditional image processing formulations of the problem usually utilize only grayscale values as features in the spatial domain~\cite{IPT14BiTDLMS}, lacking a semantic discriminability between the real targets and distractors; the resulting methods can typically handle only very salient targets of high local contrasts and not the dim ones buried in complex background.
Furthermore, algorithms exploiting such small target priors are sensitive to the hyper-parameters relevant to the image content, e.g., the sparsity control hyper-parameter $\lambda$ in low-rank methods~\cite{TGRS20TopHatReg} and the preset target size in local contrast methods~\cite{TGRS18DECM}, which fail easily in highly variable scenes with fast-changing backgrounds.

Infrared small target detection can also be modeled as a supervised \emph{machine-learning} problem, but it has long been stuck with insufficient training data due to the difficulty of collecting infrared small target images.
To this end, we recently contribute an open and high-quality dataset specially created for single-frame small infrared target detection termed ``SIRST'', which enables the training of deep networks\footnote{\url{https://github.com/YimianDai/sirst}}. 
However, with the labeled data, accurate infrared small target detection or segmentation can still be challenging for many off-the-shelf baseline networks~\cite{CVPR15FCN,MICCAI15UNet}.
The object appearance centered feature representation~\cite{TPAMI13RepresentationLearning}, which many deep networks rely on, may fail due to the scarcity of target intrinsic characteristics as well as the presence of background distractors.
Also, most convolutional networks learn high-level semantic features by gradually attenuating the feature map size, making small targets easily being overwhelmed by surrounding background features in deep layers.
This means that a specialized ``network surgery'' is essential for reliable detection performance.

In this paper, we advocate the idea of combining the feature learning capacity of deep networks and the physical mechanisms of model-driven methods into an end-to-end network to handle the intrinsic feature scarcity issue for detecting infrared small targets in a single image.
In this scheme, the backbone network extracts high-level semantic features of the input image and then a certain model-inspired module encodes them into local contrast measures.
Unlike purely data-driven methods~\cite{arXiv19TBCNet} and purely model-driven methods~\cite{TGRS17SMSL}, our approach fully makes use of both labeled data and domain knowledge. 
As a result, it solves the inaccurate modeling and hyper-parameter sensitivity problems of model-driven methods since the network learns discriminative features automatically. 
It also mitigates the minimal intrinsic feature issue of data-driven approaches by incorporating the domain knowledge of local contrast prior into deep networks.

More specifically, we propose \emph{Attentional Local Contrast Network} (ALCNet), a new model-driven deep network for single-frame infrared small target detection.
This advance stems from two key improvements over previous work.
First, by designing an acceleration scheme based on feature map cyclic shift, we modularize a local contrast measure method by Wei et al.~\cite{PR16MPCM} as a depth-wise parameterless nonlinear feature refinement layer with clear physical interpretability, which explicitly breaks the limited receptive field imposed by the local nature of convolutional kernels and encodes relatively long-range contextual interactions.
Second, to highlight and preserve the small target features, besides adjusting the network down-sampling scheme, we also exploit a \emph{bottom-up attentional modulation} (BLAM) module which encodes the smaller scale subtle details of low-level features into high-level features of deeper layers.
Finally, the cross-layer fused feature maps are used for the segmentation task.

To verify the effectiveness of the proposed ALCNet architecture, we conduct extensive ablation studies to investigate the importance of encoding local contrast prior, multi-scale local contrast measure, and bottom-up attentional modulation.
We also compare it with other state-of-the-art model-driven methods and data-driven methods on the public SIRST benchmark.  
The experimental results indicate that the proposed ALCNet achieves the best performance. 

\section{Related Works} \label{sec:related}

\textbf{Human visual system inspired methods}~\cite{TGRS13LCM} build the local contrast property into models, which is a major characteristic of these featureless infrared small targets. 
These methods depend heavily on the local contrast measure used to enhance the target and suppress the background clutters~\cite{TGRS16WLDM}.
For this purpose, different variants of local contrast measures have been proposed to deal with pixel pulse noise~\cite{GRSL14ILCM} and edges~\cite{PR16MPCM}.
However, all these methods assume a strict match between the target size and the preset nested cell or patch structure~\cite{TGRS13LCM}, as they take the mean or maximum grayscale, or entropy of each patch on the original image as hand-crafted features~\cite{TGRS18DECM}.
Such over-simplified raw features are also the root of the inaccurate modeling issue of these model-driven methods.
Although this work borrows the multi-scale patch-based contrast measure~\cite{PR16MPCM} (MPCM) as guidance, unlike the above approaches, our network does not rely on any predefined feature representation, nor do we assume that the small target has higher local contrast than any background component. 
Our network automatically learns the target feature from the labeled data, which makes it more robust in detecting dim targets in complex backgrounds.  
To the best of our knowledge, this paper is the first to modularize a conventional local contrast method and embed it into a convolutional network for infrared small target detection. 

\textbf{Small object detection} is a key challenge in generic computer vision as well because there is little signal on the object to exploit~\cite{CVPR17TinyFaces}. 
Besides carefully designing the anchor matching strategy~\cite{ICCV17S3FD} or training the network in a scale-aware scheme~\cite{NIPS18SNIPER}, encoding context that offers more evidence beyond the object extent is also highly explored as a solution to mitigate the problems arising from small objects~\cite{CVPR18EncNet,ECCV16ContextualPriming}, which is generally done by extracting and concatenating the features of an enlarged window around the objects~\cite{CVPR17TinyFaces}.
Unlike these implicit context encoding approaches, our network explicitly encodes the interaction of each element on the feature map with its distant neighbors, namely the local contrast according to some well-defined physical mechanisms~\cite{PR16MPCM}.
In addition, the scarcity of intrinsic characteristics is different between generic small objects and the infrared small target studied in this paper.
Small objects in generic vision tasks generally occupy around 1\% of image area~\cite{CVPR18SNIP}, while the pixels of an infrared small target may only take 0.1\% or less of the image area.

\textbf{Cross-layer feature fusion} has proved to be an effective approach to alleviate the scale variation issue in computer vision, which is often implemented via linear combinations such as addition or concatenation~\cite{CVPR17FPN,MICCAI15UNet}, or nonlinear top-down modulation~\cite{BMVC18PAN,SPL19SkipAttention}.
In \cref{tab:related}, we provide a brief summary of cross-layer feature integration schemes, where PWConv denotes the point-wise convolution, and Concat is the abbreviation of concatenation. 
$\mathbf{X}$ and $\mathbf{Y}$ are the feature maps of different scales. By default, $\mathbf{Y}$ is the one with a larger receptive field. 
$\mathbf{G}$ denotes the global channel attention module \cite{CVPR18SENet}, and $\mathbf{L}$ is the local channel attention module we adopt. 
However, these approaches are not designed for the task of infrared small target detection, which may fail without considering the special characteristics of this task.
Unlike the generic computer vision task, the bottleneck in infrared small target detection is how to preserve and highlight the features of dim and small targets in deep layers, rather than lacking high-level semantics in shallow layers.
To this end, our work employs a reverse bottom-up point-wise attentional modulation pathway, which we think is vital for infrared small target detection.

\setlength{\tabcolsep}{4pt}
\begin{table}[htbp]
\centering
\caption{A brief overview of cross-layer feature integration schemes. The proposed ALCNet differs in the modulation pathway direction and attention module.}
\label{tab:related}
\vspace*{-.25\baselineskip}
\small
\begin{tabular}{Sc Sc Sl} 
\toprule  
Manner               & Formulation                                                & Reference  \\
\midrule 
Addition             & $\mathbf{X} + \mathbf{Y}$                                  & \cite{CVPR17FPN} \\
Concatenation        & $\mathrm{PWConv}(\mathrm{Concat}(\mathbf{X}, \mathbf{Y}))$ & \cite{CVPR15GoogLeNet,MICCAI15UNet} \\
Refinement           & $\mathbf{X} + \mathbf{G}(\mathbf{Y}) \otimes \mathbf{Y}$   & \cite{CVPR18SENet,NIPS18GENet} \\
Top-down Modulation  & $\mathbf{G}(\mathbf{Y}) \otimes \mathbf{X} + \mathbf{Y}$   & \cite{BMVC18PAN} \\
Bottom-up Modulation & $\mathbf{X} + \mathbf{L}(\mathbf{X}) \otimes \mathbf{Y}$   & \textbf{\emph{ours}}  \\
\bottomrule
\end{tabular}
\vspace{-\baselineskip}
\end{table}


\section{Modularizing the Local Contrast Prior}
In this section, we describe how to convert a conventional patch-based local contrast measure~\cite{PR16MPCM} to a non-linear feature refinement layer which can be inserted into a network as a plug-in module.
Specially, we will deal with the challenges: 
1)~how to get rid of the patch-based paradigm of local contrast measure~\cite{TGRS13LCM} which can not fit into an end-to-end network;
2)~how to measure the local contrast on feature maps fast.
\vspace*{-.5\baselineskip}

\subsection{Dilated Local Contrast Measure}

\begin{figure}[htbp]
  \centering
  \subfloat[]{
    \includegraphics[height=.185\textwidth]{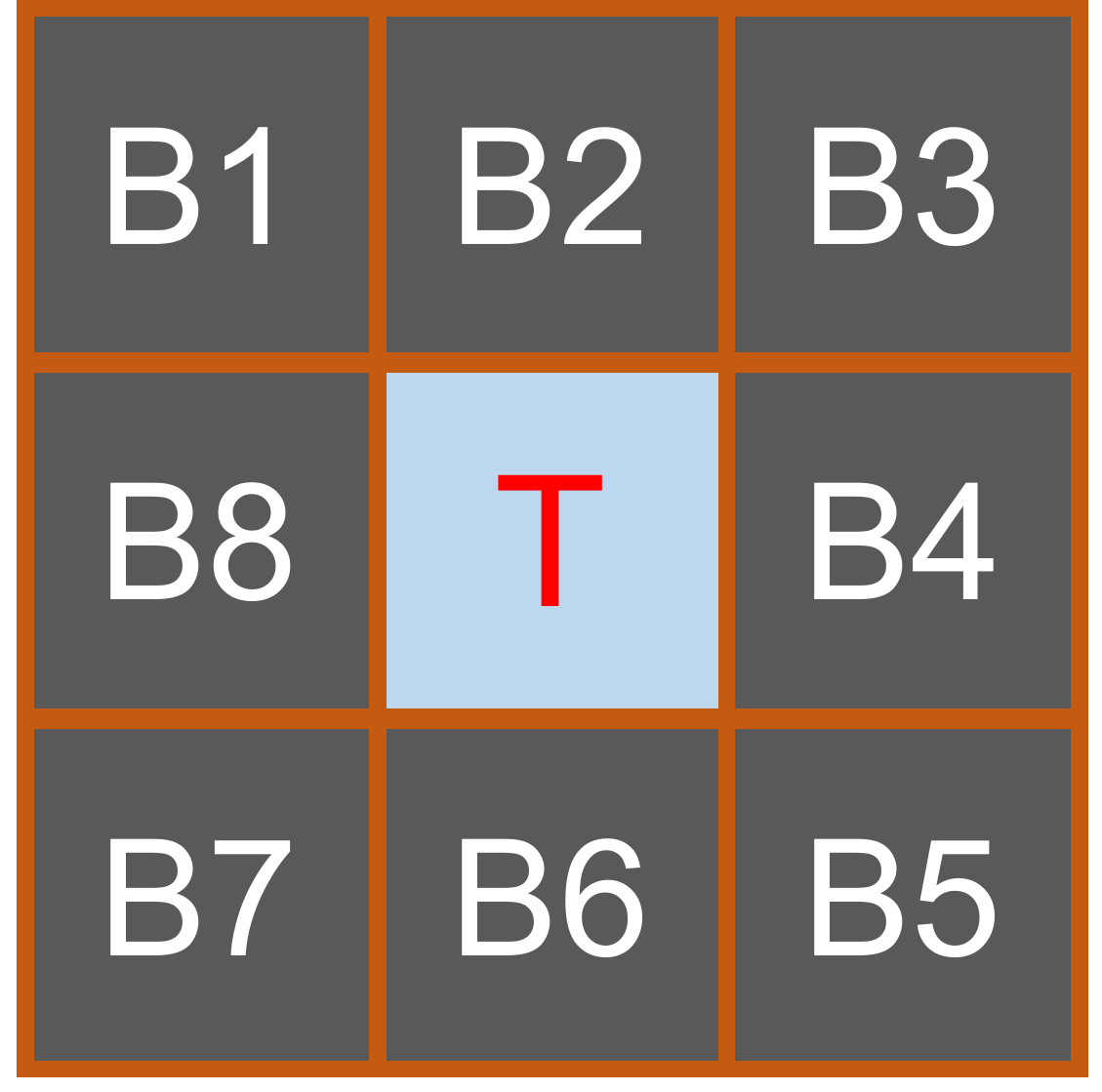}
    \label{subfig:patch}
  }\hfil
  \subfloat[]{
    \includegraphics[height=.185\textwidth]{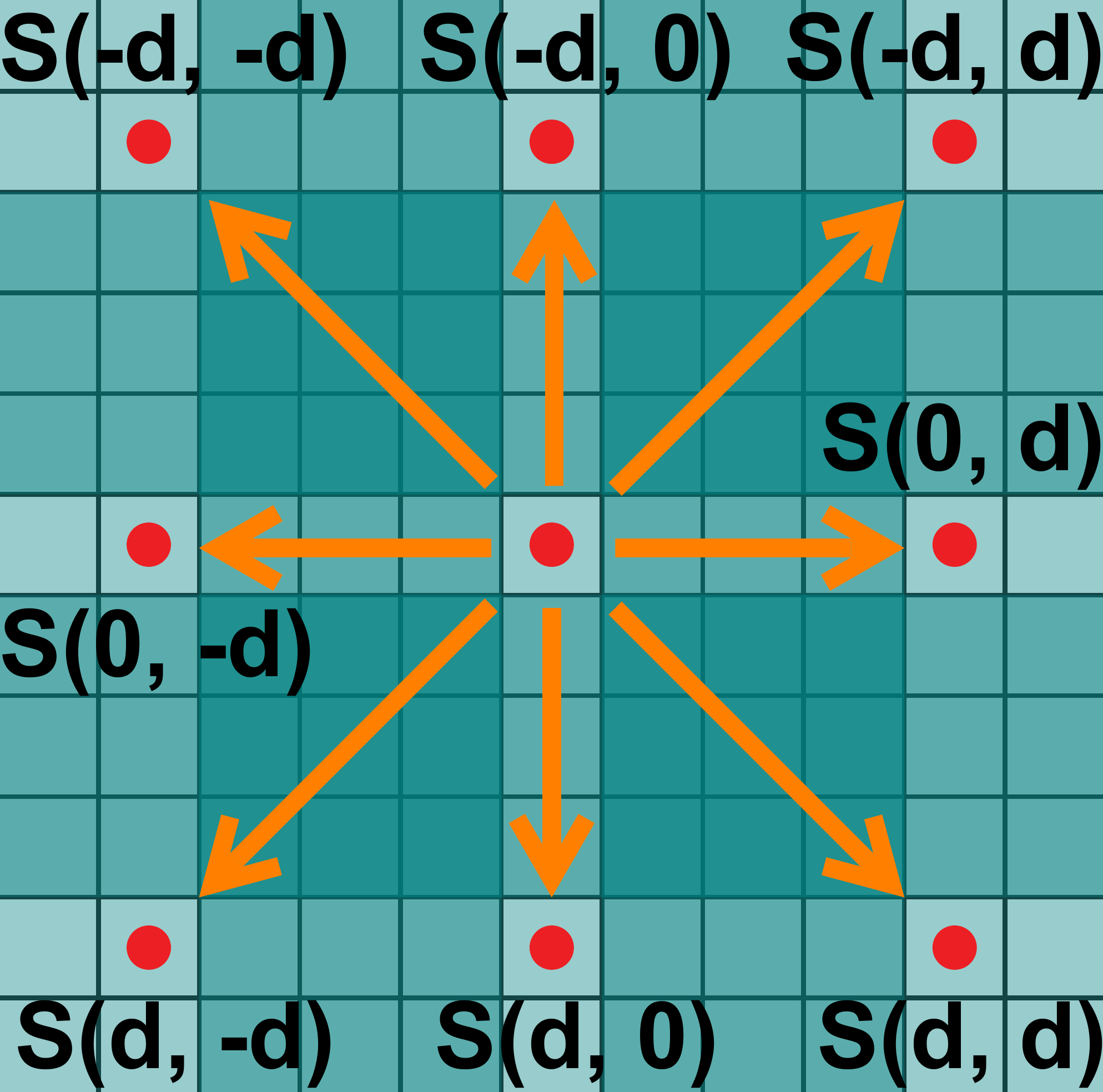}
    \label{subfig:dilation}
  }
  \caption{The transition from patch to dilation: (a)~The nested structure of patch-based contrast measure~\cite{PR16MPCM}. The patch size is equal to the filter size. Patches are strictly non-overlapped. (b)~The structure of our dilated local contrast measure, in which each point shares the overlapped receptive fields. 
  }  
  \vspace{-.5\baselineskip}
\end{figure}

In traditional local contrast measure methods~\cite{TGRS13LCM,PR16MPCM}, the patch size is both \emph{the scale of feature extraction} and \emph{the scale of local contrast measure}, imposing an explicit equality constraint on these two, as shown in \cref{subfig:patch}.
To measure the local contrast in an end-to-end network, we abandon the concept of ``patch'' and its patch mean feature representation in MPCM~\cite{PR16MPCM} .
As a replacement, we borrow the dilation rate from DeepLab series~\cite{ICLR16DilatedConv} as a hyper-parameter controlling the scale of local contrast measure, as illustrated in \cref{subfig:dilation}.
As a result, the feature scale and local contrast measure scale are decoupled in the network, which enables us to measure multi-scale local contrast on multi-scale feature maps.
\vspace*{-1.\baselineskip}

Given an intermediate feature $\mathbf{F} \in \mathbb{R}^{C \times H \times W}$, a specific position $(c, i, j)$ and a dilation rate $d$, one can write the directional local contrast in a scalar form as 
\begin{equation}
\mathbf{D}_{[c, i, j]}^{(x, y)} = \left(\mathbf{F}_{[c, i, j]} - \mathbf{F}_{[c, i-x, j-y]}\right) \cdot \left(\mathbf{F}_{[c, i, j]} - \mathbf{F}_{[c, i+x, j+y]}\right),
\label{eq:scalardirection}
\end{equation}
where $(x, y) \in \Omega = \{(-d, -d), (-d, 0), (-d, d), (0, -d)\}$ is the direction index.
Then the scalar local contrast under dilation rate $d$ can be obtained as
\begin{equation}
\mathbf{C}_{[c, i, j]}^{d} = \min_{(x, y) \in \Omega} \left\{ \mathbf{D}_{[c, i, j]}^{(x, y)} \right\}.
\vspace*{-.5\baselineskip}
\end{equation}

\subsection{Cyclic Shift Accelerating Scheme}

To fast calculate the subtraction between the center point and its neighborhoods, MPCM filters the mean image with eight pre-set kernels~\cite{PR16MPCM}. 
It should be noted that if we utilize this accelerating scheme in a depth-wise way in the network, it requires $8(3d)^2 HW$ multiplications and $8(3d)^2 HW$ additions for each feature map. 
The computational cost of the local contrast measure will grow rapidly as the dilation rate $d$ increases, which can be a bottleneck in inference speed.

\begin{figure}[htbp]
    \centering
    \subfloat{
        \includegraphics[width=0.235\textwidth]{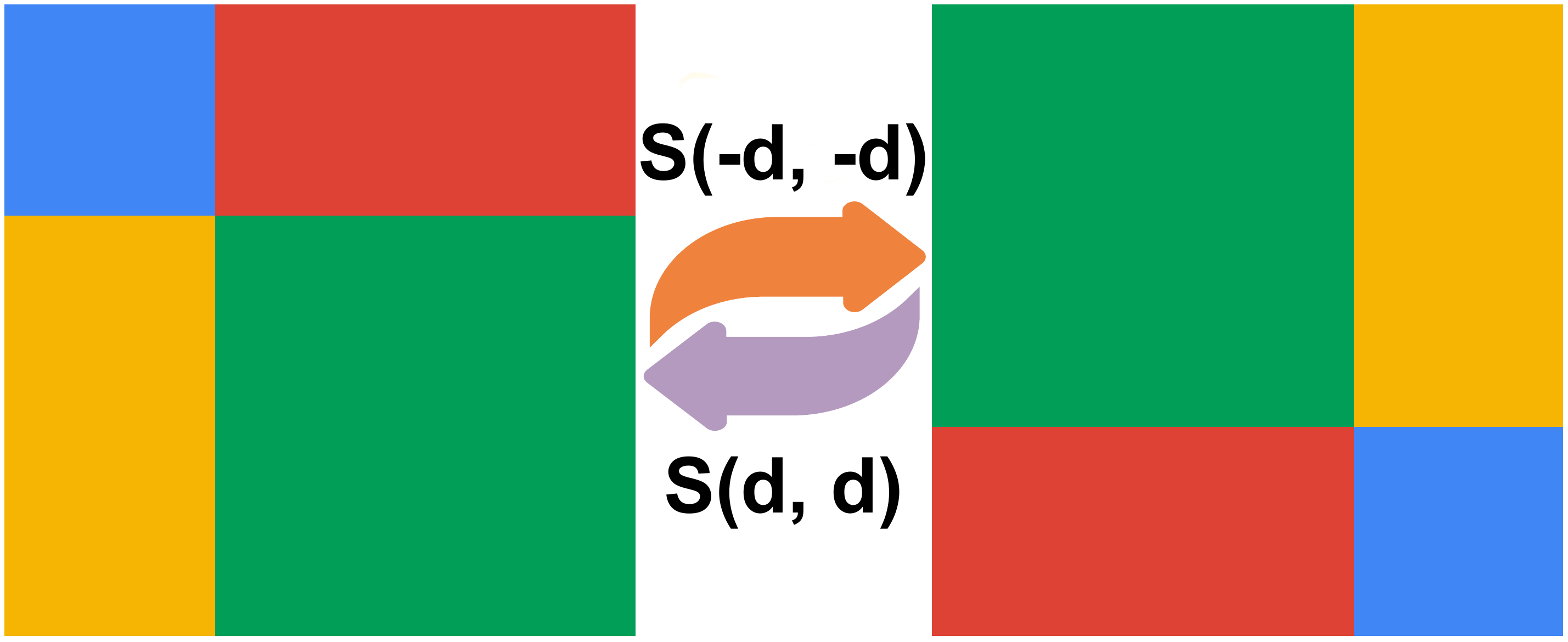}
    }\hfil
    \subfloat{
        \includegraphics[width=0.235\textwidth]{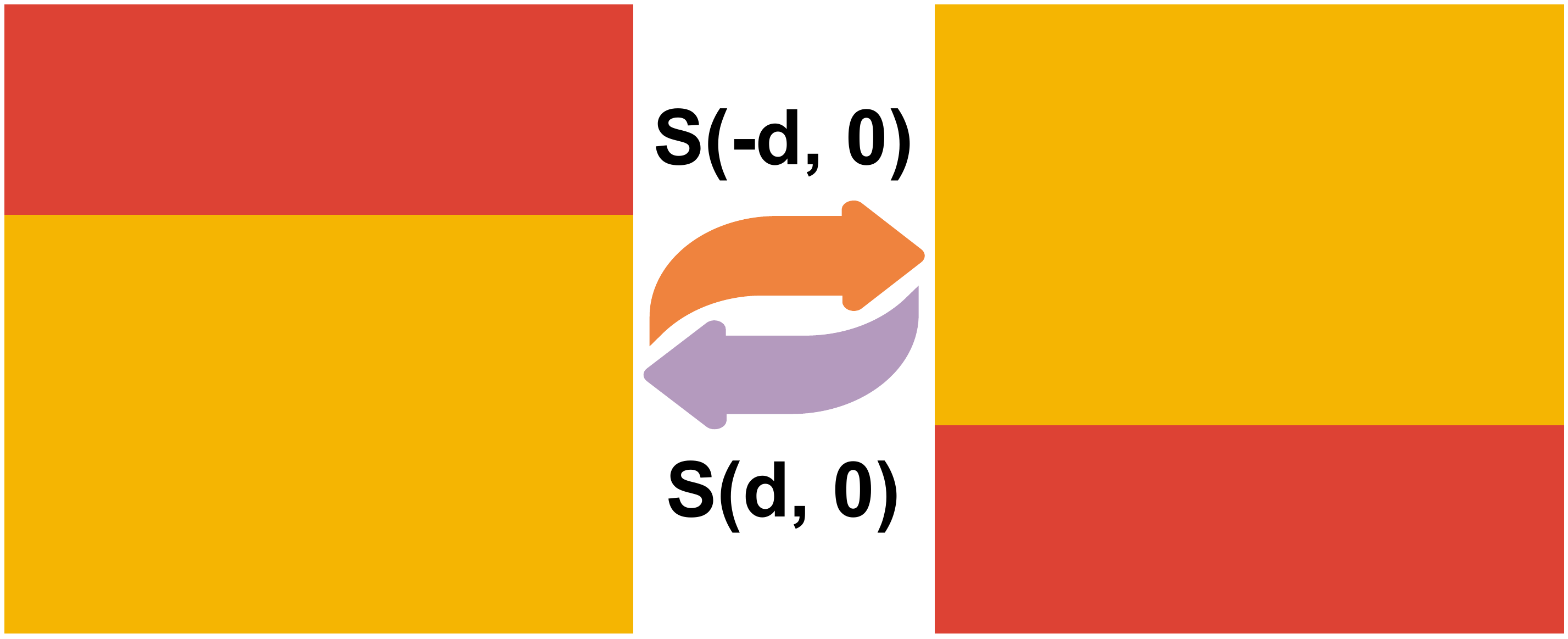}
    }
    \vspace{.25\baselineskip}
    \subfloat{
        \includegraphics[width=0.235\textwidth]{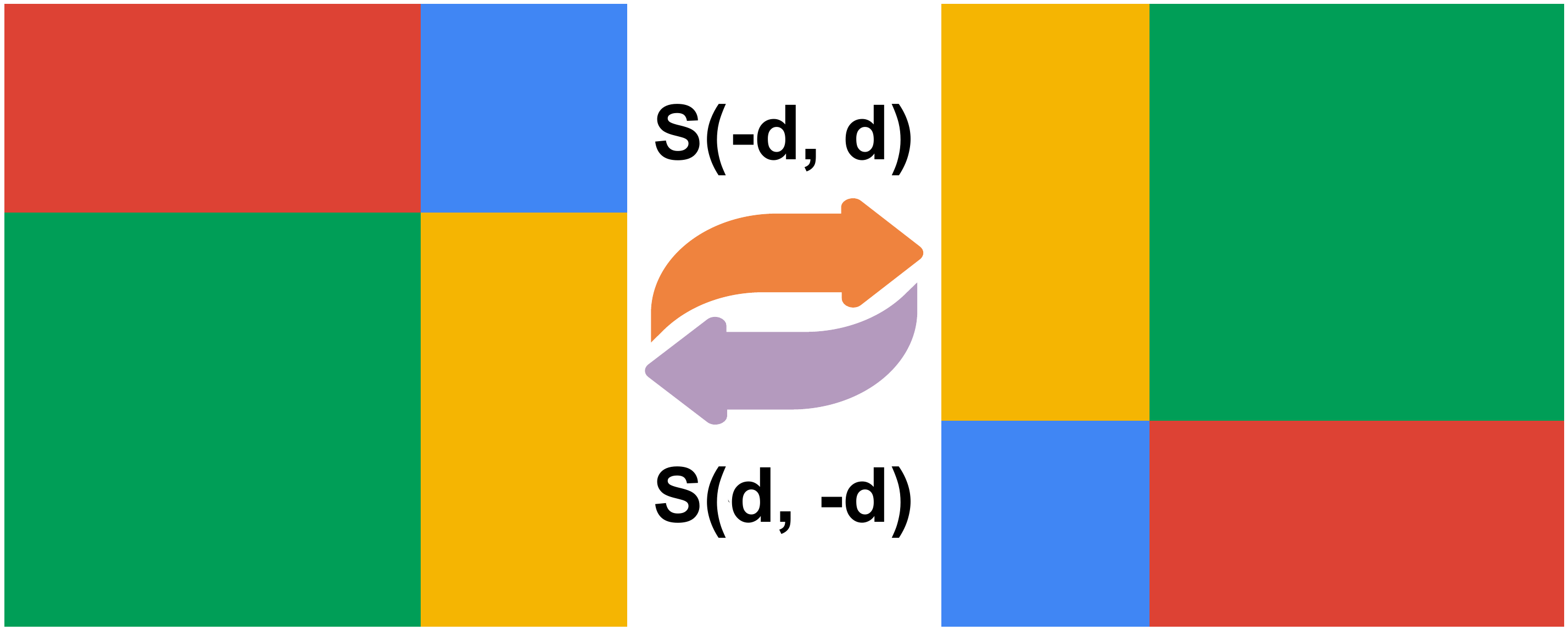}
    }\hfil
    \subfloat{
        \includegraphics[width=0.235\textwidth]{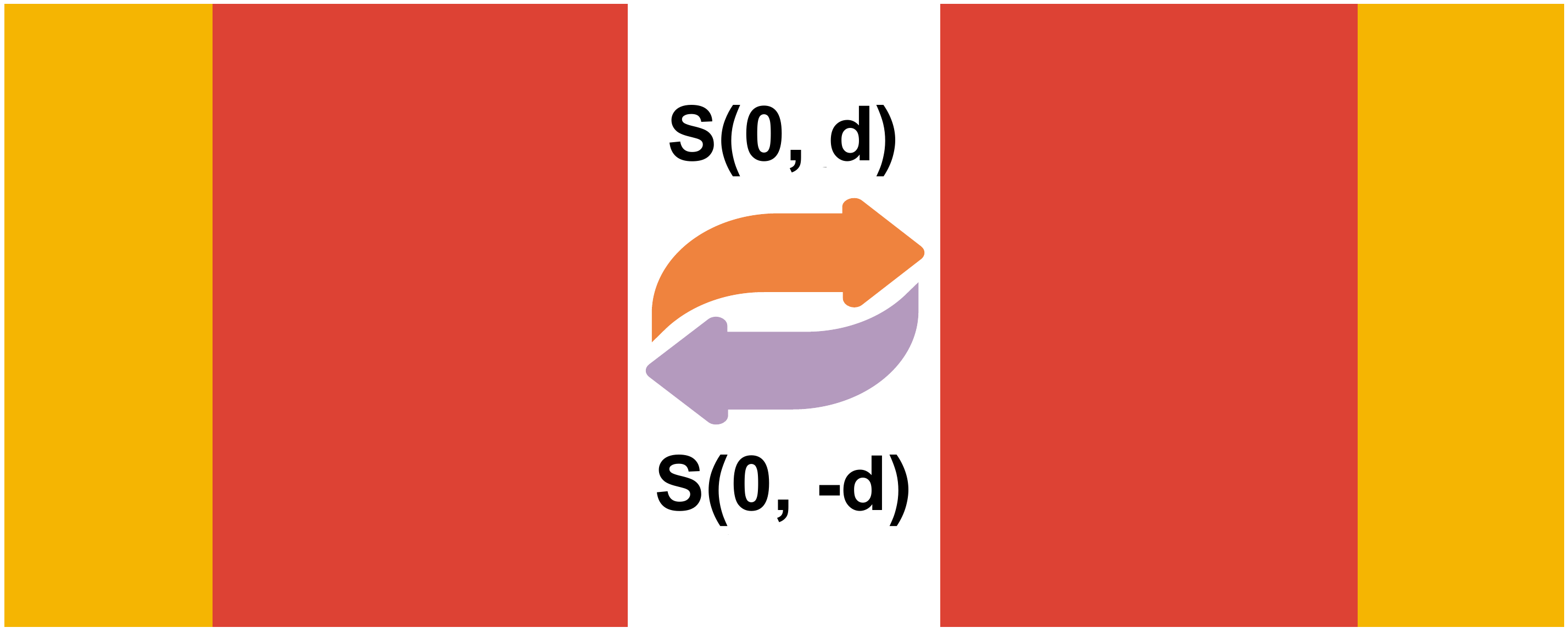}
    }        
    \caption{Illustration of the cyclic shift scheme to obtain eight neighborhood feature maps. The map at the end of the arrow denotes the original feature map, and the map at the arrowhead denotes the shifted feature map in a particular direction.}
    \label{fig:circshift}
\end{figure}

In this work, we offer another accelerating trick by hypothesizing that the feature map margins are smooth and similar. 
Actually, this is a reasonable hypothesis. 
On the one hand, the original infrared image has both strong local and non-local correlations. 
On the other hand, most background components have been suppressed by previous convolutional layers. 
Therefore, we can formulate \cref{eq:scalardirection} into a tensor form and compute it as a whole as 
\begin{equation}
\mathbf{D}^{(x, y)} = \left(\mathbf{F} - \mathbf{S}_{(x, y)}\right) \otimes \left(\mathbf{F} - \mathbf{S}_{(-x, -y)}\right),
\label{eq:contrast}
\end{equation}
where $\otimes$ is the element-wise multiplication and $\mathbf{S}_{(x, y)}$ denotes the cyclically shifted $\mathbf{F}$ in the direction $(x, y)$ with a stride $d$.
The depth-wise cyclic shift scheme is illustrated in \cref{fig:circshift}. Each arrow inside denotes the transformation from the original feature map to the shifted feature map.
The shifted feature map in the opposite direction can be obtained via reversing the cyclic shift.
With this cyclic shift trick, we can reduce the computational cost to only $8HW$ subtractions for local difference measure on each feature map.
For instance, with this trick, MPCM can be computed around 15\% faster, from 2.67 FPS to 3.07 FPS.
Then the local contrast feature map under dilation rate $d$ (DLC) can be obtained via element-wise maximum operation as
\begin{equation}
\mathrm{DLC}(\mathbf{F}, d) = \max_{(x, y) \in \Omega} \left\{ \mathbf{D}^{(x, y)} \right\}.
\vspace*{-1.5\baselineskip}
\end{equation}

\emph{Discussion.} To some extent, \cref{eq:contrast} is a special formulation of the \emph{attention mechanism} with clear physical interpretability, in which the features that match the local contrast prior are emphasized, while the rest features are suppressed. 
Under a large dilation rate $d$, the local contrast measure explicitly breaks the limited effective receptive field~\cite{NIPS16ERF} and encodes relatively long-range contextual interactions on feature maps.

\subsection{Multi-Scale Local Contrast Measure}

\begin{figure}[htbp]
  \centering
  \subfloat[MLC]{\includegraphics[height=13\baselineskip]{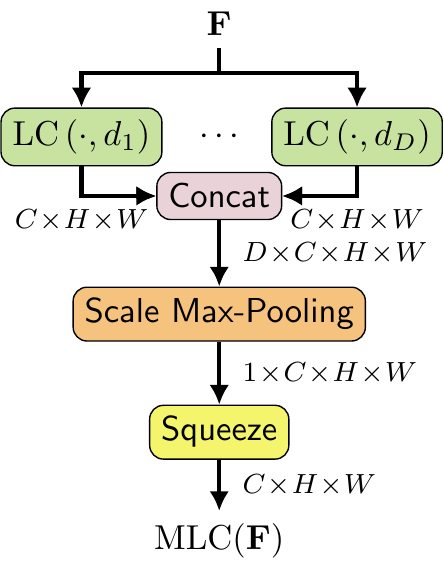}\label{fig:mlc}}
  \subfloat[BLAM]{\includegraphics[height=13\baselineskip]{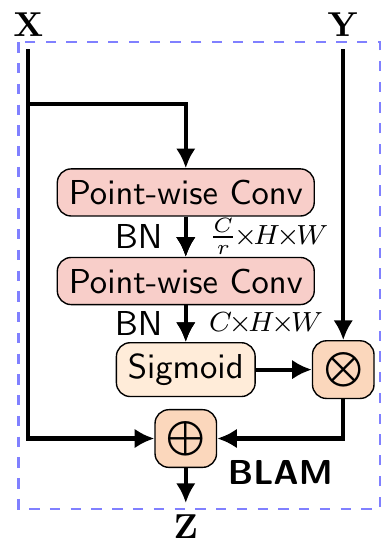}\label{fig:blam}}
  \caption{
  Illustration of the proposed modules: (a) The same-layer multi-scale local contrast (MLC) module, which embeds the local contrast prior into networks. (b) The cross-layer bottom-up local attentional modulation (BLAM) module, which embeds smaller scale details into high-level coarse feature maps.}
  \vspace{-1\baselineskip}
\end{figure}

A central issue in infrared small target detection is the scale variation of targets.
Both traditional local contrast methods and state-of-the-art convolutional network architectures share the same motivation that models should have different patch sizes or receptive fields for targets of different scales.
To remedy this issue, an intuitive way is to leverage a multi-scale local contrast measure on feature maps.
Given an intermediate feature map $\mathbf{F}$, we measure multi-scale local contrast \emph{in the same layer} $\mathrm{MLC}(\mathbf{F}) \in \mathbb{R}^{C\times H \times W}$ by applying the dilated local contrast module $\mathrm{DLC}$ with various dilation rates $\left\{d_1, d_2, \cdots, d_D\right\}$ as 
\begin{align}
  \mathrm{MLC}(\mathbf{F}) = & \mathrm{Squeeze}\left( \mathrm{SMP}\left( \mathrm{Concat}\left( \right. \right. \right. \notag \\ 
  & \left. \left. \left. \mathrm{DLC}  \left( \mathbf{F}, d_1 \right), \cdots, \mathrm{DLC} \left( \mathbf{F}, d_D \right) \right) \right) \right)  
  \label{eq:mlc}
\end{align}
where $\mathrm{SMP}$ denotes the \emph{scale max-pooling} operation that max-pools the concatenated feature tensor along the scale axis, and the $\mathrm{Squeeze}$ operation removes single-dimensional entries from the shape of the feature map. The flowchart of \cref{eq:mlc} is illustrated in \cref{fig:mlc}.
It is noteworthy that in traditional local contrast methods, it is a risky operation of taking the maximum value among different scales as the target contrast that may cause false alarms, since some background distractors are possible to have a higher contrast measure than the real targets under a preset over-simplified feature representation.
However, this problem is largely mitigated in our ALCNet because the network can \emph{learn to adjust the feature representation dynamically in an end-to-end way} according to the labeled data and loss function, thereby making sure that the real target has maximum local contrast.
\vspace*{-1.5\baselineskip}

\section{Attentional Local Contrast Network}

In this section, we describe the overall architecture and final optimization formulation of the proposed method.
Specially, we will deal with the challenges: 
1)~how to highlight the features of infrared small targets in high-level layers of a coarse representation of the input image;
2)~how to handle the class imbalance issue between small targets and the background.

\subsection{Bottom-Up Local Attentional Modulation}

A deeper network can provide better semantic features and an understanding of the context of the scene, which helps to resolve the ambiguity between the target and the background distractors. However, as the network deepens, the risk of losing the target spatial details is also increasing.
To tackle this issue, our ALCNet resorts to a cross-layer \emph{bottom-up local attentional modulation} (BLAM) module to embed low-level information into high-level coarse feature maps, as illustrated in \cref{fig:blam}. 
The local channel attention mechanism $\mathbf{L}$ locally aggregates channel feature context for each spatial position individually as
\begin{equation}
\mathbf{L}(\mathbf{X}) = \sigma \left( \mathcal{B} \left(\mathrm{PWConv}_2 \left( \delta\left( \mathcal{B} \left(\mathrm{PWConv}_1 (\mathbf{X}) \right)\right)\right)\right) \right)
\end{equation}
where $\mathrm{PWConv}$, $\sigma$, $\delta$, and $\mathcal{B}$ denote the Point-Wise Convolution \cite{ICLR14NiN}, Sigmoid function, Rectified Linear Unit (ReLU) \cite{ICML10ReLU}, and Batch Normalization (BN) \cite{ICML15BN}, respectively.
The kernel sizes of $\mathrm{PWConv}_1$ and $\mathrm{PWConv}_2$ are $\frac{C}{4} \times C \times 1 \times 1$ and $C \times \frac{C}{4} \times 1 \times 1$, respectively, consisting of a bottleneck structure.
It is noteworthy that the attentional weight map $\mathbf{L}(\mathbf{X}) \in \mathbb{R}^{C \times H \times W}$ has the same shape as the input feature maps and can thus be used to highlight the subtle details in an element-wise manner---spatially and across channels.
Therefore, the BLAM module is able to be dynamically aware of the subtle details of small infrared targets.

The motivation of the BLAM module is to embed smaller scale details into high-level coarse feature maps, which is achieved as a dynamically weighted modulation of the high-level features under the guidance of low-level features.
Given $\mathbf{X}$ as the low-level feature and $\mathbf{Y}$ the high-level feature, the cross-layer fused feature $\mathbf{Z}^{\prime} \in \mathbb{R}^{C \times H \times W}$ can be obtained via the bottom-up local attentional module as
\begin{equation}
  \mathbf{Z} = \mathbf{X} + \mathbf{L}(\mathbf{X}) \otimes \mathbf{Y}
\end{equation}
where $\otimes$ denotes the element-wise multiplication. 
In the context of multi-scale local contrast measure, by replacing $\mathbf{X}$ and $\mathbf{Y}$ with $\mathrm{MLC}(\mathbf{X})$ and $\mathrm{MLC}(\mathbf{Y})$, we can obtain the fused multi-scale local contrast feature map $\mathbf{Z}^{\prime} \in \mathbb{R}^{C \times H \times W}$ via
\begin{align}
  \mathbf{Z}^{\prime} & = \mathrm{MLC}(\mathbf{X}) \uplus \mathrm{MLC}(\mathbf{Y}) \notag\\
  & = \mathrm{MLC}(\mathbf{X}) + \mathbf{L}\left(\mathrm{MLC}(\mathbf{X})\right) \otimes \mathrm{MLC}(\mathbf{Y}),
\end{align}
where $\uplus$ denotes the cross-layer feature fusion via BLAM.

\subsection{Network Architecture}

A high-resolution prediction map is vital for detecting infrared small targets due to their small sizes.
To preserve the small target, we use a modified ResNet-20 \cite{ECCV16ResNetV2} as the backbone network to extract the feature maps. 
To unlike conventional approaches that downsample the image 32 times, the backbone in our ALCNet only subsamples the image twice at stage2\_1 and stage3\_1 with a stride of 2, as shown in \cref{Tab:Backbone}. 
To stack images of different sizes into a batch, each image is resized to $512 \times 512$ and randomly cropped to $480 \times 480$ during training.

\newcommand{\blockb}[2]{$
\begin{bmatrix}
\begin{array}{l}
    3 \times 3 \mathrm{~conv}, #1 \\
    3 \times 3 \mathrm{~conv}, #1 \\    
\end{array}
\end{bmatrix} \times #2$}

\newcolumntype{x}[1]{>\centering p{#1pt}}
\newcommand{\ft}[1]{\fontsize{#1pt}{1em}\selectfont}
\renewcommand\arraystretch{1.25}
\setlength{\tabcolsep}{3pt}

\begin{table}[htbp]
\caption{
Backbone architecture. 
We scale the model by depth (the block number $b$ in each stage) to study the relationship between the performance and network depth.
When $b = 3$, it is the standard ResNet-20 backbone \cite{ECCV16ResNetV2}.
}
\label{Tab:Backbone}
\centering
\begin{tabular}{Sc Sc Sc}
\toprule  
Stage & Output & ResBlock \\ 
\midrule 
Conv-1  & 480 $\times$ 480 & \hspace{-2.25em} $3\times 3~\mathrm{conv}, 16$ \\ 
Stage-1 & 480 $\times$ 480 & \blockb{16}{b} \\
Stage-2 & 240 $\times$ 240 & \blockb{32}{b} \\
Stage-3 & 120 $\times$ 120 & \blockb{64}{b} \\
\bottomrule
\end{tabular}
\vspace{-.5\baselineskip}
\end{table}

\begin{figure}[htbp]
  \centering
  \includegraphics[width=0.48\textwidth]{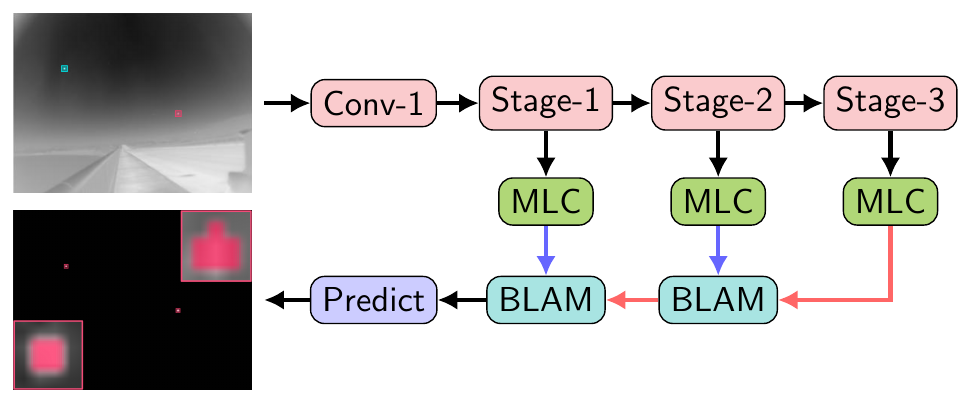}
  \caption{Architectures of the proposed ALCNet, which incorporates same-layer multi-scale local contrast (MLC) modules and cross-layer bottom-up local attentional feature modulation (BLAM) modules into a feature pyramid network. The blue line and red line represent the channel number transformation and the upsampling operator, respectively.}
  \label{fig:alcnet}
  \vspace{-1.5\baselineskip}
\end{figure}

With the BLAM module, the multi-scale local contrast feature maps of different stages can be further fused in a smart way, which recovers the full spatial resolution at the network output by iteratively fusing coarse, high-layer feature maps with fine, low layer feature maps as
\begin{align}
  \mathrm{M^2LC}(f) = & \uplus \left( \mathrm{MLC}(\mathbf{F}^{(1)}), \uplus \left( \cdots, \right. \right. \notag\\
 & \left. \left. \uplus \left( \mathrm{MLC}(\mathbf{F}^{(L-1)}), \mathrm{MLC}(\mathbf{F}^{(L)}) \right) \right) \right),
 \label{eq:m2lc}
\end{align}
where $\mathrm{M^2LC}(f) \in \mathbb{R}^{C\times H \times W}$ is the final local contrast feature maps given an infrared image $f$ and $\uplus$ denotes the BLAM module. 
It should be noted that for the sake of simplicity, \cref{eq:m2lc} omits $1 \times 1$ convolutions which are used to adjust the number of filters in the proposed ALCNet. 
Finally, the two-stage multi-scale local contrast feature maps $\mathrm{M^2LC}(f)$ are used to predict small infrared targets.
The whole proposed network that conducts two-stage multi-scale local contrast measurement is proposed illustrated as \cref{fig:alcnet}.
\vspace*{-.5\baselineskip}

\subsection{Problem Formulation and Optimization}

To handle the class imbalance issue between infrared small target and the background, we adopt the Soft-IoU loss function~\cite{SoftIoU} for this highly unbalanced segmentation task, which is defined as follows:
\begin{equation}
\ell_{\text {soft-IoU}}(p, y)= \frac{\sum_{i,j} p_{i,j} \cdot y_{i,j}}{\sum_{i,j} p_{i,j} + y_{i,j} - p_{i,j} \cdot y_{i,j}},
\end{equation}
where $p = \sigma(\mathrm{M^2LC}(f, \Theta)) \in \mathbb{R}^{H \times W}$ is the prediction score map and $y \in \mathbb{R}^{H \times W}$ is the labeled mask, given an infrared image $f$. 
$\Theta$ denotes the weights of the proposed ALCNet.
Given $N$ training samples, $\Theta$ will be learned during training by minimizing the total loss as
\begin{equation}
\Theta = \arg\min_{\Theta}\  \sum_{n = 1}^{N} \ell_{\text {soft-IoU}}(\sigma(\mathrm{M^2LC}(f, \Theta))_n, y_n),
\end{equation}
For optimization, we adopt AdaGrad~\cite{COLT10Adagrad} as an optimizer with a learning rate of $0.1$ and the strategy described by He~\etal~\cite{ICCV15PReLU} for weight initialization, a total of 400 epochs, weight decay of $10^{-4}$, and a batch size of 10. 

\section{\textbf{Experiments}}

To analyze the potential of the proposed ALCNet, we compare it with state-of-the-art baselines on the public SIRST dataset. 
We also conduct a comprehensive ablation study to investigate the effectiveness of the design of the ALCNet and its behavior under different parameter and computational budgets. 
In particular, the following questions will be investigated in our experimental evaluation:
\begin{compactenum}
  \item Q1: Our key insight is to embed the local contrast measure into convolution networks as a parameterless non-linear feature extraction layer. Given the same parameter budget, we investigate the question of how the proposed dilated local contrast module helps learn more discriminative features for infrared small targets, see \cref{subsubsec:prior}.
  \item Q2: From another perspective, the proposed ALCNet can be viewed as a learnable and dynamic multi-scale local contrast feature measure. We will examine the impact of multi-scale local contrast feature fusion in the same layer and across layers, respectively, as well as the importance of bottom-up local attentional modulation (see \cref{subsubsec:scale}).
  \item Q3: Generally, multi-scale feature fusion is done by integrating high-level semantic information into low-level features in a top-down manner, but our ALCNet utilizes a reverse bottom-up local attentional modulation instead. 
  In our study (see \cref{subsubsec:attention}), we investigate the question of how important the bottom-up modulation and the local feature context are for infrared small targets.
  \item Q4: Finally, we will analyze how the proposed ALCNet compares to other state-of-the-art model-driven or data-driven methods, see \cref{subsec:sota}.    
\end{compactenum}
\vspace*{-.5\baselineskip}

\subsection{Experimental Settings}
Before investigating the questions Q1~-~Q4, we first introduce our experimental settings in detail.

\subsubsection{Dataset}
\begin{figure*}[htbp]
  \centering
  \includegraphics[width=0.975\textwidth]{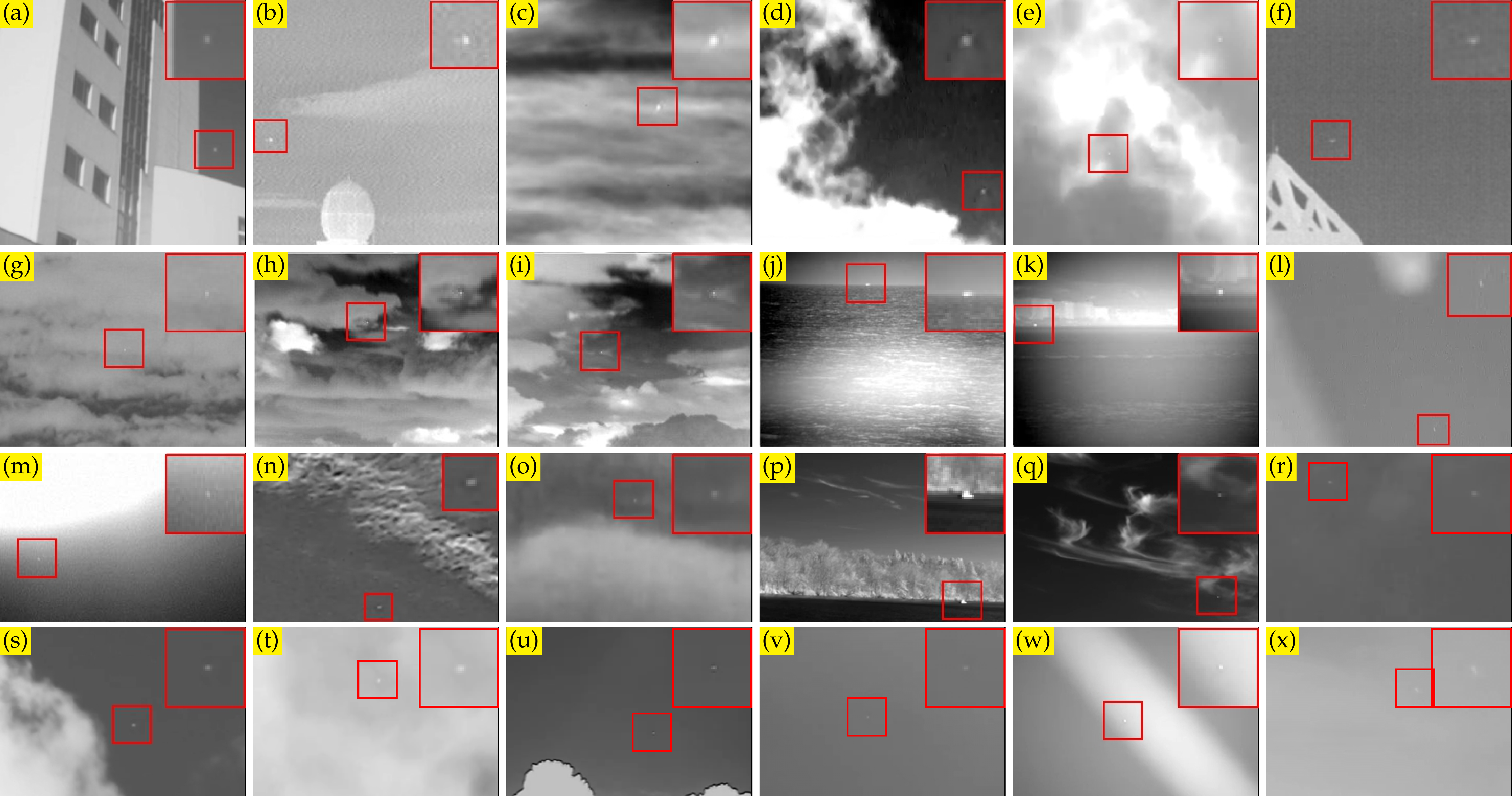}   
  \caption{The representative infrared images from the SIRST dataset with various backgrounds, which excludes many trivial cases. 
  For better visualization, the demarcated area is enlarged, which is better to be seen by zooming on a computer screen.
  }  
  \label{fig:gallery}
  \vspace{-1\baselineskip}
\end{figure*}

For experimental evaluation, we resort to our public SIRST dataset, which is the largest open dataset for single-frame infrared small target detection to the best of our knowledge. 
It contains 427 representative images and 480 instances of different scenarios from hundreds of real-world videos and is roughly split into approximately 50\% train, 20\% validation, and 30\% test. 
\cref{fig:gallery} illustrates some representative images from the SIRST dataset, from which we can see that many infrared small targets are extremely dim and buried in complex backgrounds with heavy clutter. 
In addition, only 35\% targets in the dataset contain the brightest pixel in the image. 
Therefore, the methods purely based on the target saliency assumption or merely thresholding on the raw image would not work well.

\subsubsection{Implementation Details}
For data-driven methods, 
\emph{Feature pyramid networks} (FPN) \cite{CVPR17FPN}, \emph{selective kernel networks}~\cite{CVPR19SKNet} style FPN (SK-FPN), \emph{FPN with global attention upsampling} (GAU-FPN)~\cite{BMVC18PAN}, and \emph{TBC-Net}~\cite{arXiv19TBCNet} are selected for comparison.
These methods share the same loss function, optimizer, and other hyper-parameters as the proposed ALCNet.
For non-learning model-driven methods, we choose \emph{stable multi-subspace learning} (SMSL)~\cite{TGRS17SMSL}, \emph{facet kernel and random walker} (FKRW)~\cite{TGRS19FKRW}, \emph{multi-scale patch-based contrast measure} (MPCM)~\cite{PR16MPCM}, \emph{infrared patch-image model} (IPI)~\cite{TIP13IPI}, \emph{non-negative IPI model via partial sum minimization of singular values} (NIPPS)~\cite{IPT17NIPPS}, and \emph{reweighted infrared patch-tensor model} (RIPT)~\cite{JSTARS17RIPT} for comparison. 
Their detailed hyper-parameter settings are listed in \cref{tab:params}, which are determined by an exhaustive search on the trainval set of the SIRST dataset.

\setlength{\tabcolsep}{4pt}
\begin{table*}[htbp]
\caption{Detailed hyper-parameter settings of model-driven methods for comparison.}
\label{tab:params}
\vspace*{-.5\baselineskip}
\centering
\small
\begin{tabular}{Sl Sl} 
\toprule
Methods & Hyper-parameter settings \\
\midrule
MPCM \cite{PR16MPCM} & $N=1,3,...,9$ \\
FKRW \cite{TGRS19FKRW} & $K=4$, $p=6$, $\beta=200$, window size:$ 11\times11$ \\
SMSL \cite{TGRS17SMSL} & Patch size: 50$\times$50, $\lambda = \frac{2\times L}{\sqrt{\min{(m,n)}}}$, $L=2.0$, threshold factor: $k=1$ \\
IPI \cite{TIP13IPI} & Patch size: 50$\times$50, stride: 10, $\lambda {\rm{ = }}L{\rm{/min(m,n}}{{\rm{)}}^{1/2}}$,$L=4.5$, threshold factor: $k=10$, $\varepsilon {\rm{ = 1}}{{\rm{0}}^{{\rm{ - 7}}}}$ \\
NIPPS \cite{IPT17NIPPS} &  Patch size: 50$\times$50, stride: 10, $\lambda = \frac{L}{\sqrt{\min{(m,n)}}}$, $L=2.0$, energy constraint ratio: $r=0.11$, threshold factor: $k=10$ \\
RIPT \cite{JSTARS17RIPT} &  Patch size: 50$\times$50, stride: 10, $\lambda = \frac{L}{\sqrt{\min{(I,J,P)}}}$, $L=0.001$, $h=0.1$, $\epsilon$=0.01, $\varepsilon = 10^{-7}$, threshold factor:$k=10$ \\
\bottomrule
\end{tabular}
\vspace{-1\baselineskip}
\end{table*}

\subsubsection{Evaluation Metrics}

Besides the \emph{intersection over union} (IoU) metric and \emph{receiver operating characteristic} (ROC) curve, we also choose the normalized IoU (nIoU) to evaluate the proposed ALCNet. 
nIoU is specially designed for the SIRST dataset as a more balanced metric between model-driven and data-driven methods, which is defined as 
\begin{equation}
  \mathrm{nIoU} = \frac{1}{N} \sum_{i}^{N} \frac{\mathrm{TP}[i]}{ \mathrm{T}[i] + \mathrm{P}[i] - \mathrm{TP}[i]}
\end{equation}
where, $\mathrm{TP}$, $\mathrm{T}$, and $\mathrm{P}$ denote the true positive, true and positive, respectively.
Unlike conventional filtering or target-background separation approaches, the proposed ALCNet outputs binary decisions. Therefore, traditional evaluation metrics for background suppression, including local signal to noise ratio gain, background suppression factor, and signal to clutter ratio gain are not suitable here. 

\subsection{Ablation Study}\label{subsec:ablation}

\begin{figure}[htbp]
    \vspace*{-1.\baselineskip}
    \centering
    \subfloat[]{
        \includegraphics[height=12\baselineskip]{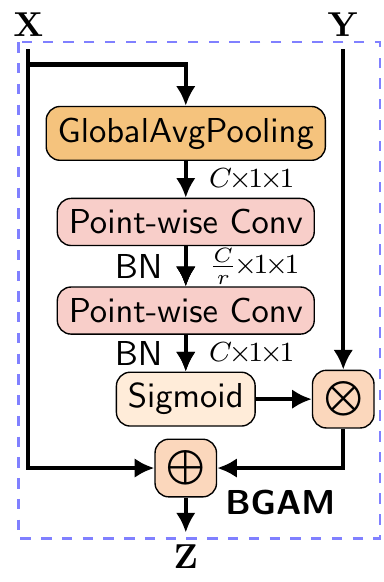}
        \label{subfig:BGA}
    }\hfil
    \subfloat[]{
        \includegraphics[height=12\baselineskip]{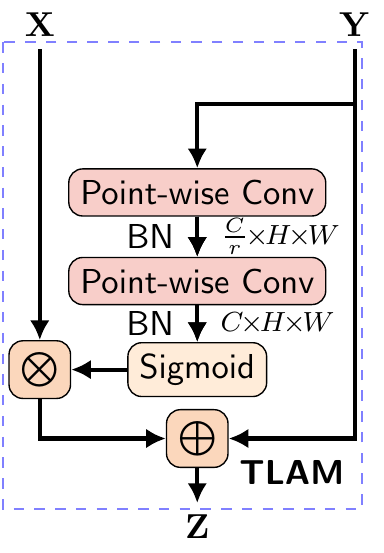}
        \label{subfig:TLA}
    }
    \caption{Architectures for the ablation study: (a) Bottom-up global attentional modulation (BGAM) module and (b) Top-down local attentional modulation (TLAM) module. 
    }
    \vspace{-1.5\baselineskip}
\end{figure}

We start by investigating the questions Q1~-~Q3 raised above. 
To better understand the proposed ALCNet, we consider several competitors constructed by removing or replacing specific parts of ALCNet.
In \cref{tab:ablationarch}, we illustrate these ablation study architectures according to their first-stage same-layer and second-stage cross-layer feature extraction schemes.
``Plain'' means that no local contrast module is used.
The \emph{bottom-up global attention module} (BGAM) and \emph{top-down local attention module} (TLAM) are illustrated in \cref{subfig:BGA} and \cref{subfig:TLA}.

\setlength{\tabcolsep}{4pt}
\begin{table}[htbp]
\begin{center}
\caption{Illustration of architectures of their different same-layer multi-scale local contrast feature extraction and cross-layer feature fusion schemes in the ablation study.}
\label{tab:ablationarch}
\vspace*{-.5\baselineskip}
\small
\begin{tabular}{Sc Sc Sc} 
\toprule  
Same-Layer & Cross-Layer & Architecture \\
\midrule 
Plain & None & PlainFCN  \\
Plain & Skip Connection via Addition & FPN \\
DLC & Skip Connection via Addition & DLC-FPN \\
MLC & Skip Connection via Addition & MLC-FPN \\
\addlinespace
MLC & Skip Connection via Maximum & Max-FPN \\
MLC & Top-down Local Attention & TLA-FPN \\
MLC & Bottom-up Global Attention & BGA-FPN \\
MLC & Bottom-up Local Attention & ALCNet (ours) \\
\bottomrule
\end{tabular}
\end{center}
\vspace{-2\baselineskip}
\end{table}

\subsubsection{Impact of Local Contrast Prior (Q1)}\label{subsubsec:prior}

\begin{figure}[htbp]
    \centering


    \includegraphics[width=0.35\textwidth]{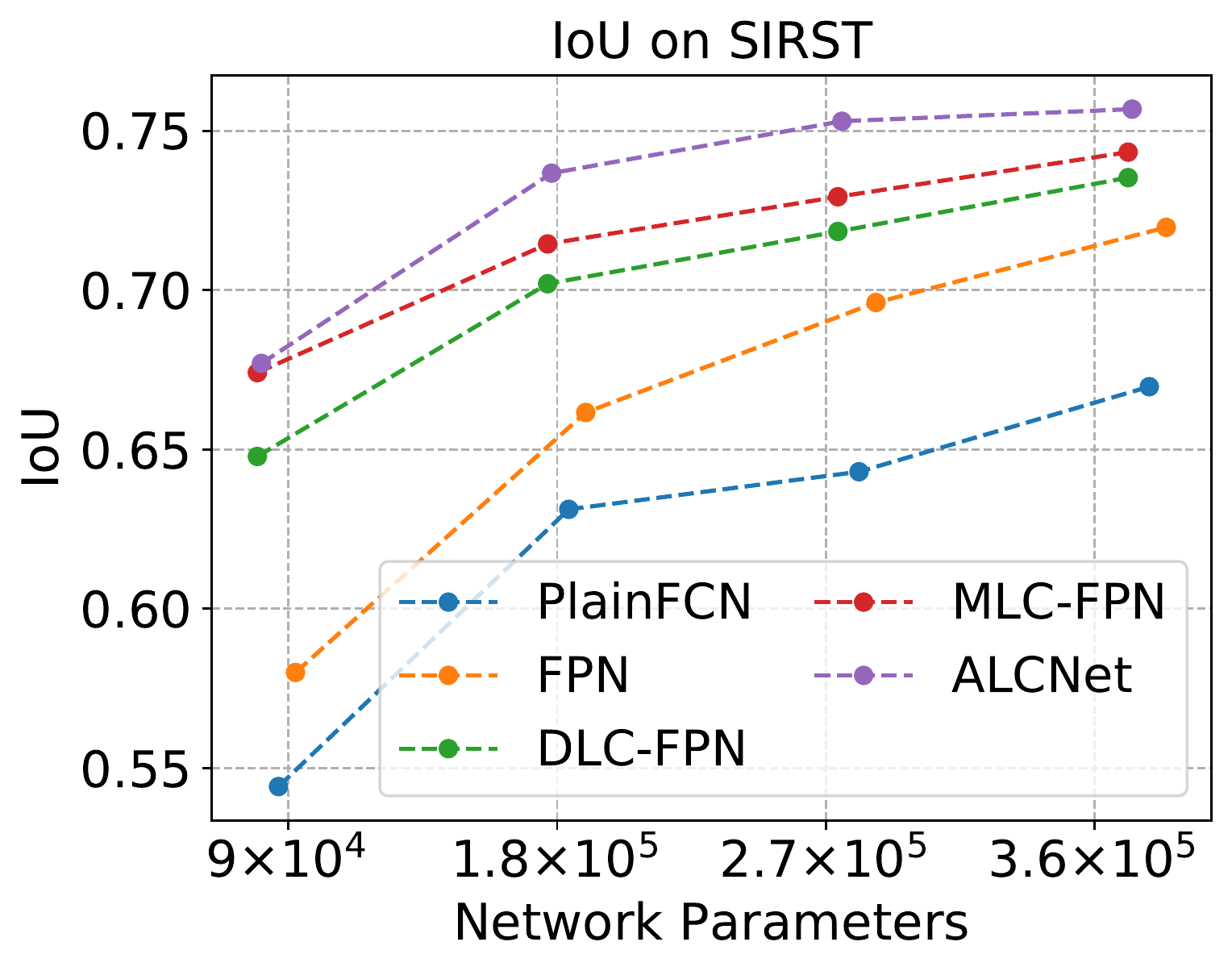}\\
    \includegraphics[width=0.35\textwidth]{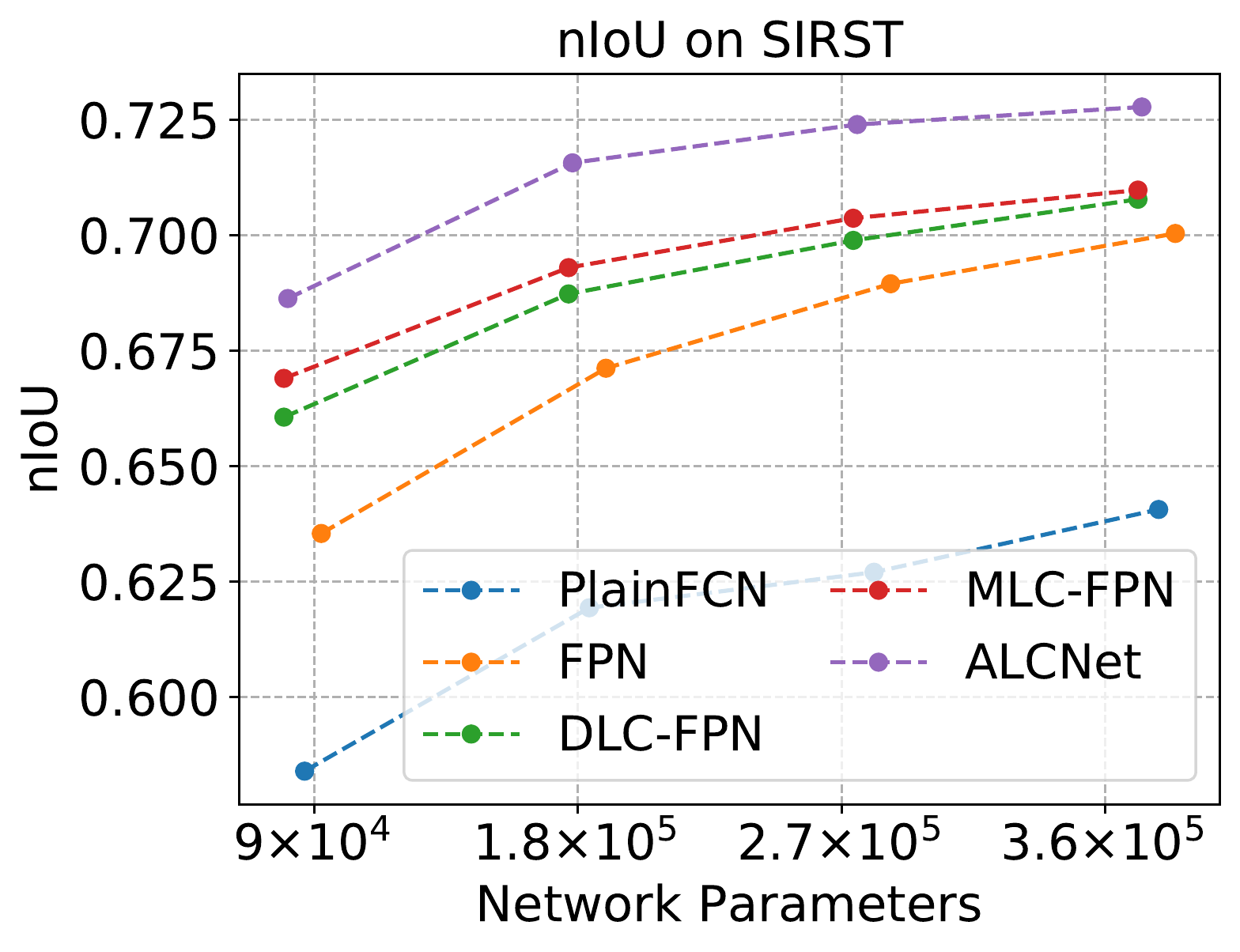}
    \caption{
    Performance comparison of ablation architectures. 
    The comparison among FPN, DLC-FPN, and MLC-FPN suggests that incorporating local contrast prior helps and a multi-scale measure in the same layer can further boost the performance.
    The results comparing PlainFCN and FPN suggest that cross-layer feature fusion is of vital importance.
    Further, the results comparing MLC-FPN and ALC-Net suggest that one should pay more attention to the cross-layer feature fusion, and a more sophisticated scheme yields consistent performance gains.}
    \label{fig:ablation}
    \vspace{-2.\baselineskip}
\end{figure}

We start by comparing the FPN and DLC-FPN that embeds a dilated local contrast (DLC) module before cross-layer fusion. 
Note that, in FPN, we add a $3 \times 3$ convolution as post-processing to improve its performance, while DLC-FPN does not have such post-processing, and the DLC module does not induce additional parameters. 
Therefore, the FPN has a bit more parameters than DLC-FPN when $b$ is the same.
\cref{fig:ablation} presents their comparison on IoU and nIoU given a gradually increased network depth. 
The dilation rate of DLC-FPN is $13$.
It can be seen that, compared with FPN, the performance of the DLC-FPN is consistently and significantly better. 
Especially the performance of DLC-FPN when $b=3$ is approximately the same as FPN ($b=4$). 
This result suggests that incorporating the local contrast prior into deep networks helps alleviate the minimal intrinsic feature issue.

This performance gain can be explained from two perspectives.
On the one hand, the local contrast module transforms the network from appearance-based recognition to local contrast-based recognition, which helps suppress more background clutter with the domain knowledge. 
On the other hand, it can also be viewed as a particular spatial attention module with clear physical interpretability, which encodes relatively long-range contextual interactions in a depth-wise manner.

\subsubsection{Impact of Multi-scale Local Contrast Integration (Q2)}\label{subsubsec:scale}
\begin{figure}[htbp]
  \centering
  \vspace{-.5\baselineskip}
  \includegraphics[width=0.35\textwidth]{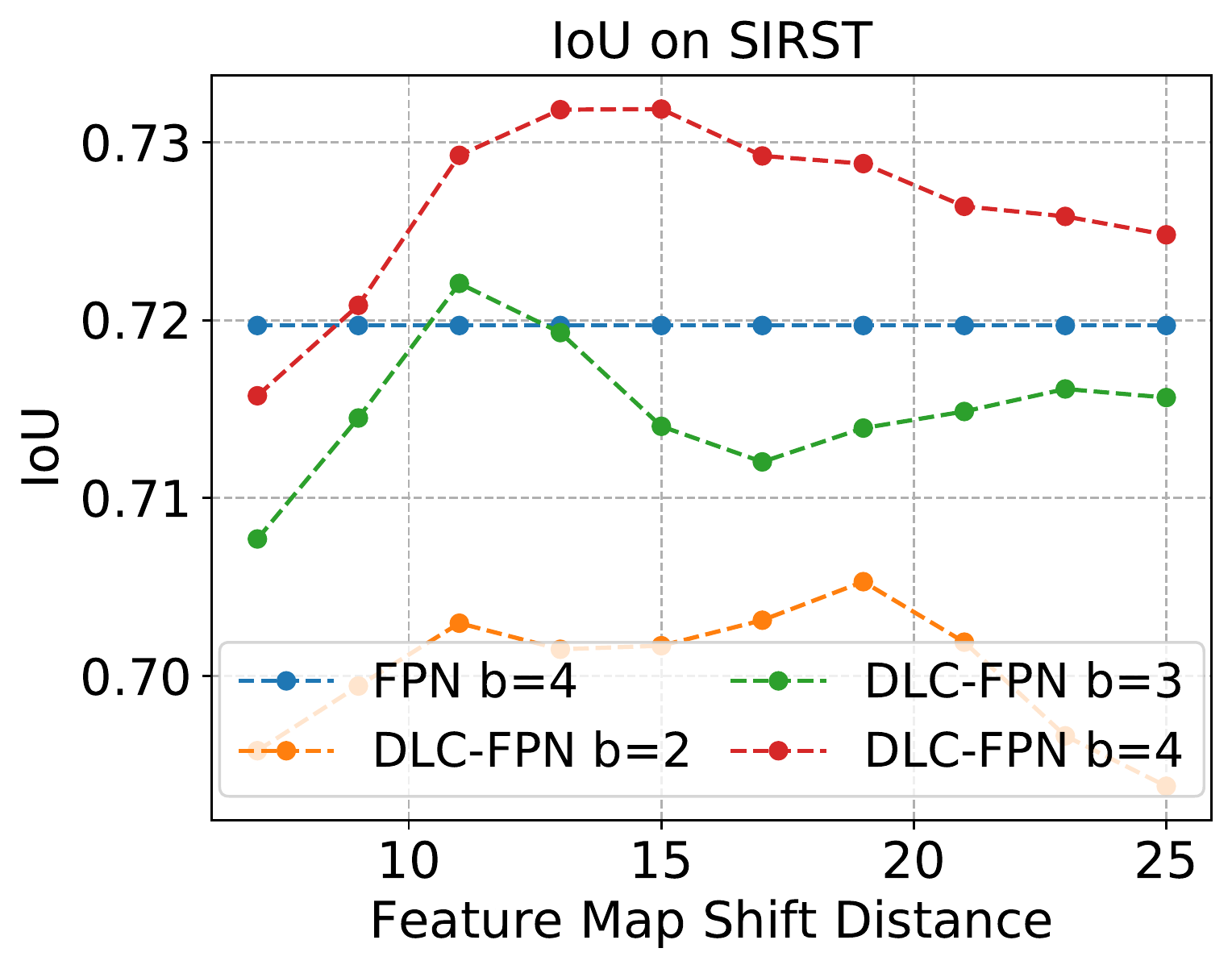}\\
  \includegraphics[width=0.35\textwidth]{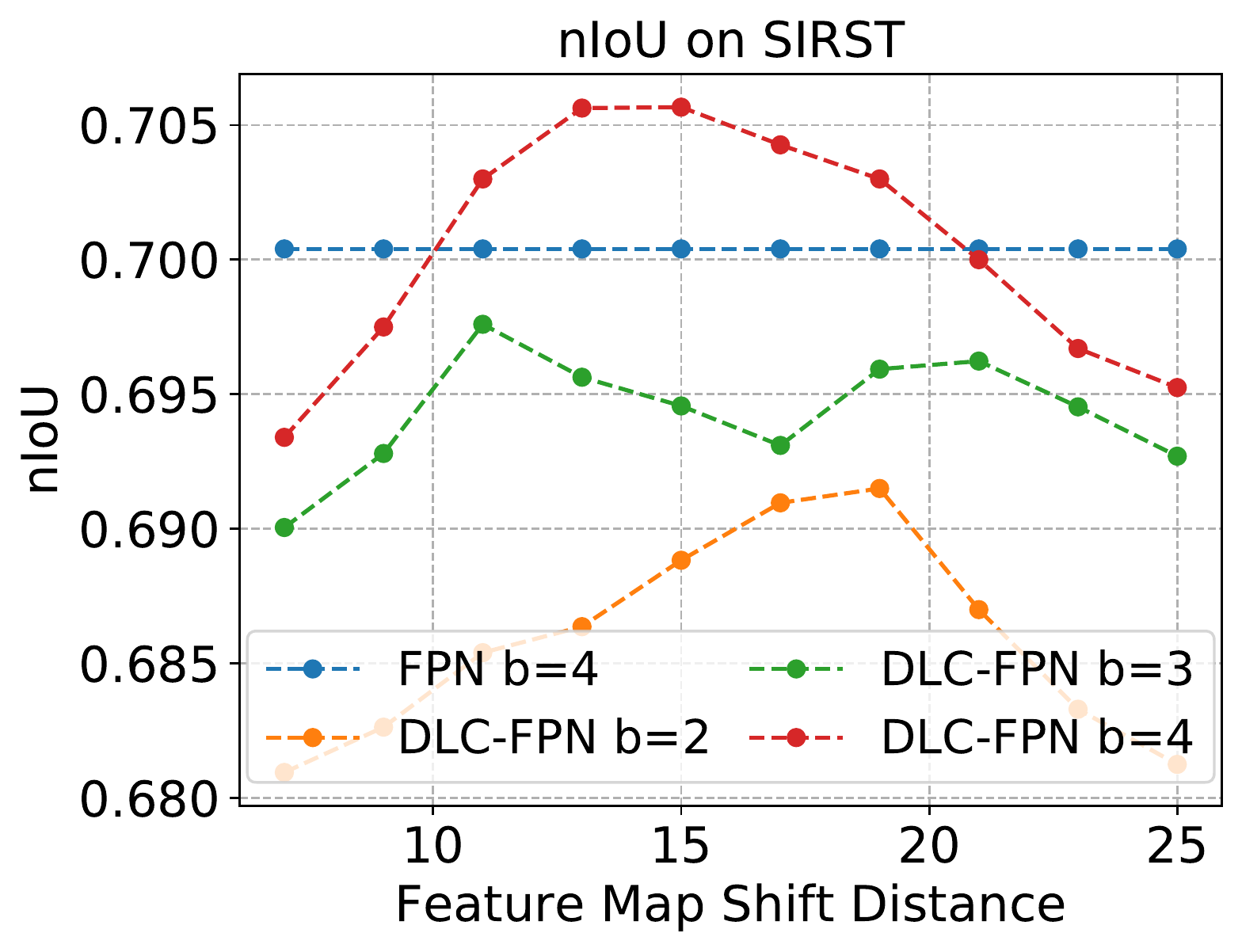}
  \caption{IoU and nIoU performance of DLC-FPN with varying dilation rates and network depths, which suggest that a single-scale local contrast measure is sensitive to the dilated value.
  }
  \label{fig:scale}
  \vspace{-.5\baselineskip}
\end{figure}

Next, we investigate the importance of the multi-scale local contrast measure in our ALCNet.
In \cref{fig:scale}, we provide the performance of DLC-FPN with various dilation rates and network depth and comparison with FPN ($b=4$). 
It can be seen that, just like its non-learning counterparts, our dilated local contrast network is sensitive to the dilation rate, which is a hyper-parameter. 
Generally, DLC-FPN performs better than FPN. 
However, with a lousy dilation rate, DLC-FPN is possible to perform worse than FPN, especially on the nIoU metric.

To solve this issue, we adopt the multi-scale local contrast (MLC) measure on the same layer feature maps.
\cref{fig:ablation} presents the comparison between single-scale DLC-FPN and multi-scale MLC-FPN given a gradually increased network depth. 
The dilation rates of MLC-FPN are $13$ and $17$.
It can be seen that by covering multiple dilation rates, the performance of MLC-FPN is consistently better than DLC-FPN. 
The MLC measure enables the network to aggregate multi-scale spatial information in the same layer, which greatly improves the robustness against the target scale variation.
It should be noted that the gap between DLC-FPN and MLC-FPN is not so significant because we select the best dilation rate for DLC-FPN. 

The importance of the cross-layer local contrast feature fusion is also shown in \cref{fig:ablation}.
PlainFCN that did not fuse the feature maps of different layers via skip connections is significantly worse.
The results suggest that cross-layer feature integration is of vital importance for infrared small targets. 
Further, ALCNet that replaces the simple element-wise addition with our specially designed BLAM module performs consistently better than the rest competitors.
Especially compared with the second-best MLC-FPN ($b=4$), our ALCNet can perform similarly on the IoU metric and even better on the nIoU metric with only around $50\%$ parameters.
The results suggest that one should pay attention to the cross-layer feature fusion for infrared small targets and that more sophisticated fusion mechanisms hold the potential to consistently yield better results.
We believe the reason behind the performance gain brought by the cross-layer feature fusion is that the low-level features offer detailed information for the accurate localization of small targets, and the high-level features help solve the ambiguous cases with its semantic information.


\subsubsection{Impact of Cross-layer Fusion Manners (Q3)}\label{subsubsec:attention}
Next, we also investigate and compare our bottom-up local attention modulation (BLAM) module with two other ablation modules. The first one is the bottom-up global attentional modulation (BGAM) module, which aggregates the global contextual information by adding a global averaging pooling at the beginning of the local channel attention module, as shown in \cref{subfig:BGA}. 
The second one that reversed the ALCNet’s modulation direction from bottom-up to top-down is the top-down local attentional modulation (TLAM) module shown in \cref{subfig:TLA}.
\cref{tab:ablation} provides the results, from which it can be seen that:
1)~The performance of BGAM-FPN is not as good as TLAM-FPN and ALCNet, which suggests that for infrared small targets, given the same additional budget for parameters and computation costs, one should aggregate the fine local information as guidance instead of the global contextual information.
2)~The difference between TLAM-FPN and ALCNet is that TLAM-FPN embeds the high-level semantic information into low-level features in a top-down manner. In contrast, the proposed ALCNet reverses this modulation direction by using the low-level feature as guidance to refine the high-level feature maps.
The results are strong support for designing bottom-up attentional modulation pathways for infrared small targets.
It is because unlike the semantic segmentation task in generic vision datasets, the localization error occupies most of the overall error. Therefore, compared with top-down semantic guidance, bottom-up detailed information is more helpful for accurate segmentation.

\setlength{\tabcolsep}{3pt}
\begin{table*}[ht]
\begin{center}
\begin{threeparttable}
\caption{Comparison of different cross-layer feature fusion schemes on the SIRST dataset.
The results comparing BGA-FPN and ALCNet suggest that the local contextual aggregation is vital for infrared small targets.
The comparison between TLA-FPN and ALCNet suggests that given the same computational and parameter budget, the bottom-up modulation performs better than the top-down one.}
\label{tab:ablation}
\small
\begin{tabular}{Sc Sc Sc Sc Sc Sc Sc Sc Sc Sc Sc Sc} 
\toprule  
\multirow{2}{*}[-1ex]{\shortstack[c]{Contextual\\Scale}} & \multirow{2}{*}[-1ex]{\shortstack[c]{Modulation\\Direction}} & \multirow{2}{*}[-0.ex]{Formulation} & \multirow{2}{*}[-0.ex]{\shortstack[c]{Architecture}} & \multicolumn{4}{c}{IoU} & \multicolumn{4}{c}{nIoU} \\
\cmidrule{5-8} \cmidrule{9-12} 
     &      &                   &             & $b=1$ & $b=2$ & $b=3$ & $b=4$ & $b=1$ & $b=2$ & $b=3$ & $b=4$ \\
\midrule 
\multirow{2}{*}{None} & \multirow{2}{*}{None} & $\mathbf{X} + \mathbf{Y}$ & FPN      & 0.674 & 0.713 & 0.729 & 0.744 & 0.669 & 0.691 & 0.702 & 0.710 \\
 &  & $\max(\mathbf{X}, \mathbf{Y})$ & Max-FPN & 0.665 & 0.713 & 0.722 & 0.734 & 0.674 & 0.698 & 0.706 & 0.712 \\
Global & Bottom-Up & $\mathbf{X} + \mathbf{G}(\mathbf{X}) \otimes \mathbf{Y}$ & BGA-FPN & 0.676 & 0.714 & 0.731 & 0.736 & 0.679 & 0.698 & 0.704 & 0.711 \\
\multirow{2}{*}{Local} & Top-Down & $\mathbf{L}(\mathbf{X}) \otimes \mathbf{X} + \mathbf{Y}$ & TLA-FPN & \textbf{0.688} &  0.729 & 0.750  & 0.753 & \textbf{0.688} & 0.708 & 0.722 & 0.718 \\
& Bottom-Up & $\mathbf{X} + \mathbf{L}(\mathbf{X}) \otimes \mathbf{Y}$ & ALCNet & 0.677 & \textbf{0.737} & \textbf{0.753} & \textbf{0.757} & 0.686 & \textbf{0.716} & \textbf{0.724} & \textbf{0.728} \\
\bottomrule
\end{tabular}
\end{threeparttable}
\end{center}
\vspace{-1\baselineskip}
\end{table*}


\subsection{Comparison to State-of-the-art Approaches}\label{subsec:sota}
\begin{figure*}[htbp]
  \vspace{-1.\baselineskip}
    \centering
    \subfloat[]{
        \includegraphics[height=10.4\baselineskip]{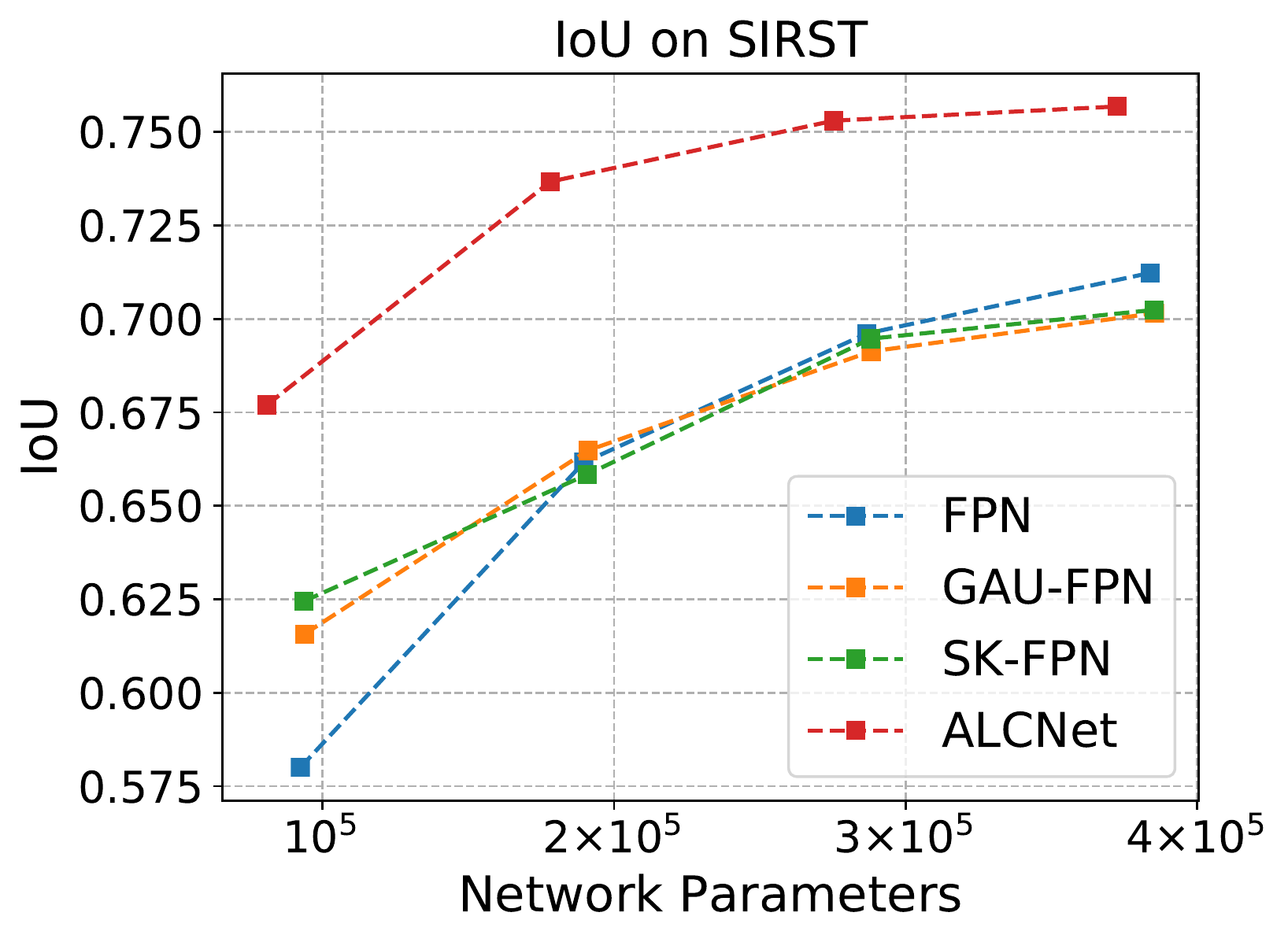}\label{subfig:iou}
    }\hfil
    \subfloat[]{
        \includegraphics[height=10.4\baselineskip]{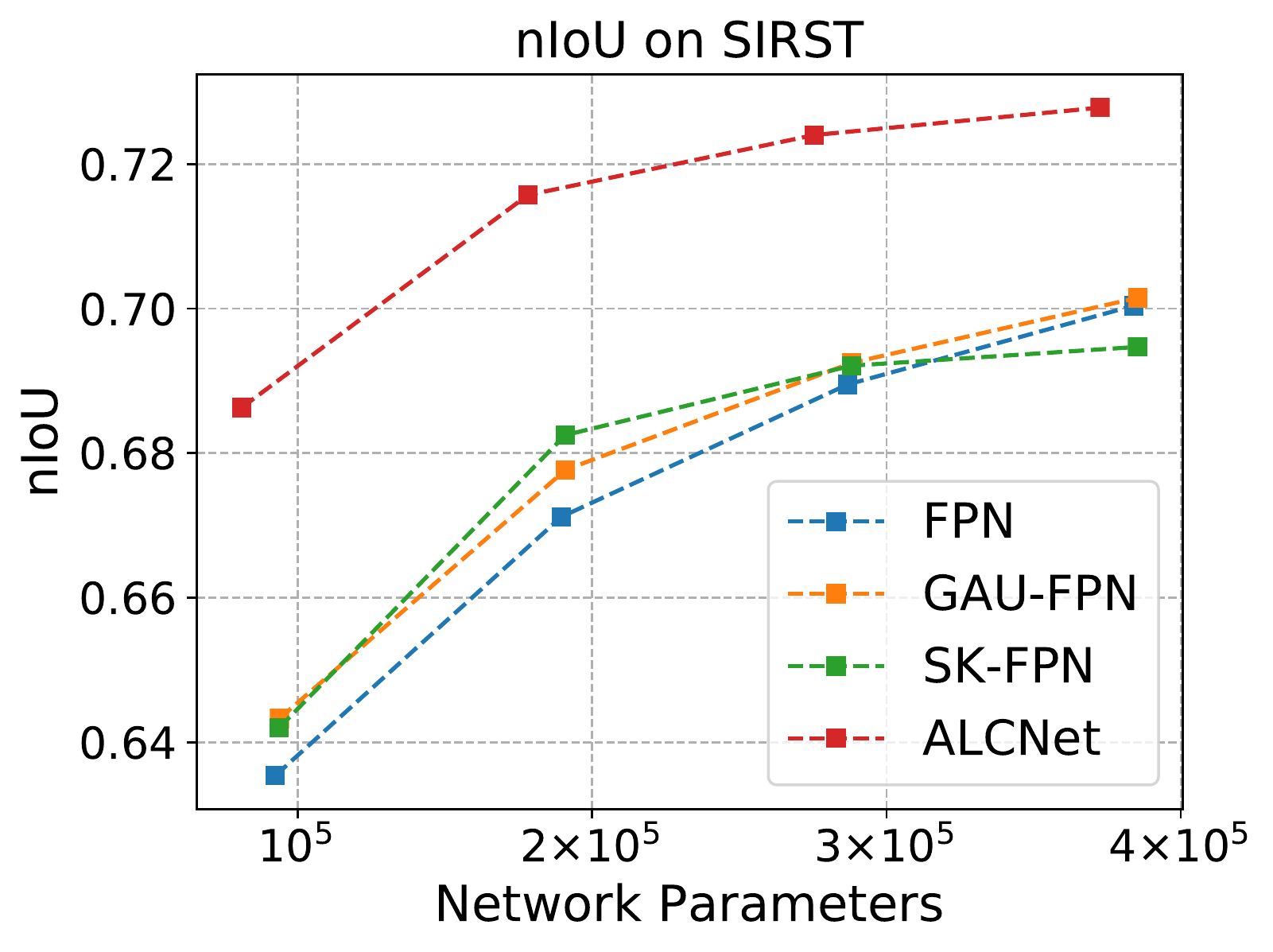}\label{subfig:niou}
    }\hfil
    \subfloat[]{
        \includegraphics[height=10.4\baselineskip]{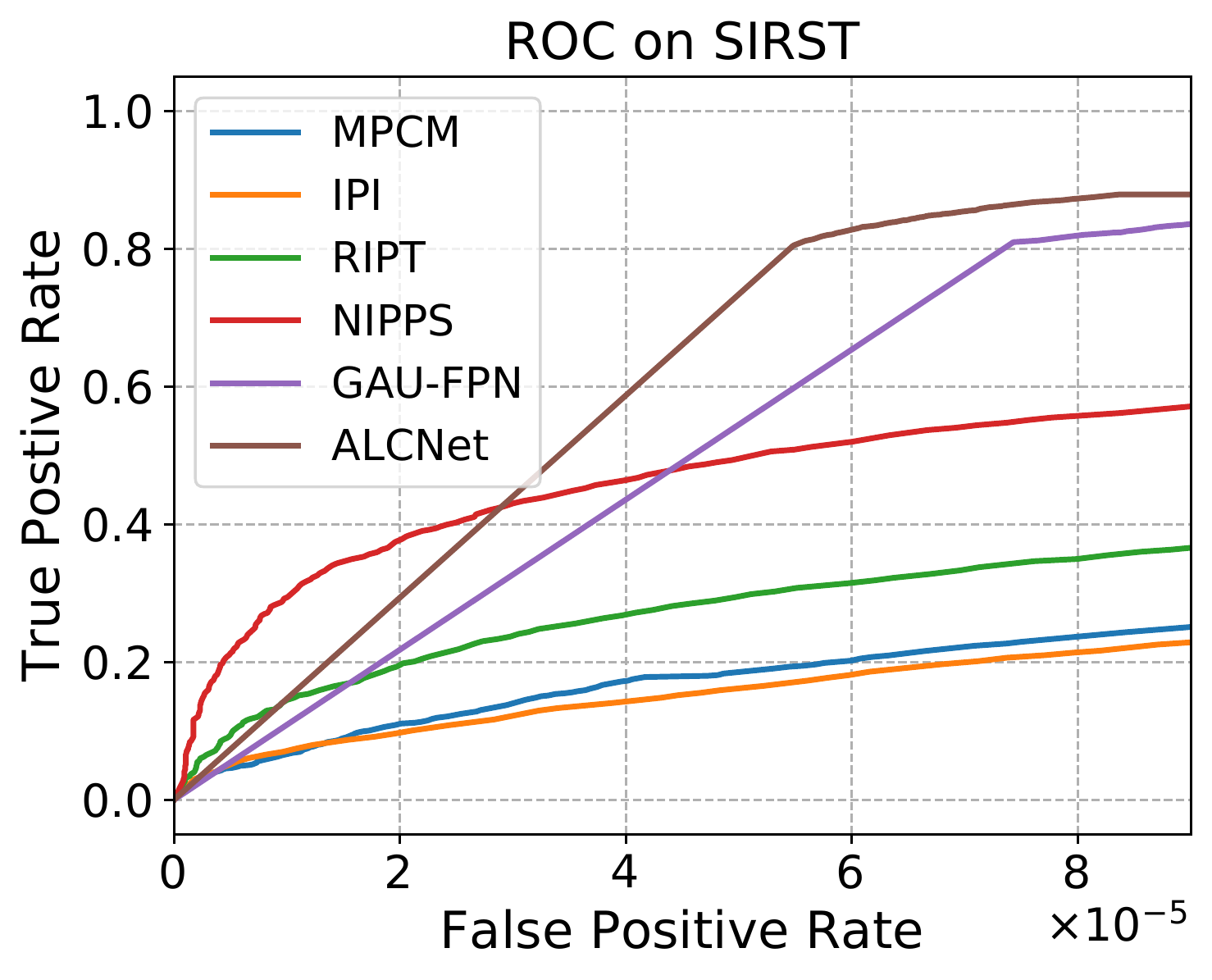}\label{subfig:roc}
    }    \\[-1.ex]
    \caption{The predictive performance comparison with other state-of-the-art methods: (a) and (b) Comparison with other state-of-the-art baseline networks on IoU and nIoU with a gradual increase of network depth. (c) Comparison with both data-driven methods and model-driven methods on ROC. The proposed ALCNet consistently yields the best performance.}
    \vspace{-.5\baselineskip}
\end{figure*}

Finally, we address question Q4 by comparing our ALCNet with several model-driven methods and other state-of-the-art network competitors. 
First, we compare the proposed ALCNet with other deep convolutional networks, namely, FPN \cite{CVPR17FPN}, SK-FPN \cite{CVPR19SKNet}, and GAU-FPN \cite{BMVC18PAN}, on the SIRST dataset, given a gradual increase of the depths of the networks. The results are provided in \cref{subfig:iou} and \cref{subfig:niou}.
It can be seen that: (a)~The proposed ALCNet achieves a significantly better performance for all experimental settings, which demonstrates its effectiveness compared to other baselines.
These results reaffirm that one can obtain better infrared small detection performance by incorporating local contrast domain knowledge and by dynamically modulating the high-level features with guidance from low-level feature maps. 
(b)~Even with a half parameter number ($b=2$),  ALCNet still performs better than these baseline networks with $b=4$. The results suggest that by utilizing the local contrast prior and paying attention to the cross-layer feature fusion, one can obtain a more efficient convolutional network that yields better performance with fewer layers or parameters per network.

Next, we compare our proposed ALCNet with the baseline networks and other state-of-the-art non-learning model-driven methods on the IoU and nIoU metrics in \cref{Tab:IoU} as well as the computational time.
We also validate our ALCNet on the ROC metric. The results are provided in \cref{subfig:roc}.  
It can be seen that
1)~All convolutional networks perform better than the non-learning model-driven methods, which shows that learning from the data offers a promising way leading to better performance for infrared small detection.
2)~Our ALCNet achieves the best among learning and non-learning approaches, showing the effectiveness of the proposed architecture.
Compared to GAU-FPN, the proposed ALCNet achieves a higher detection rate and a lower false-alarm rate simultaneously.
3)~It should be noted that to keep the comparison fair, the average inference time per sample of every method including both model-driven methods and deep networks is evaluated as each sample at a time with the same CPU.
It can be seen that once finishing training, the inference of deep networks is much faster than conventional saliency detection or low-rank plus sparse decomposition methods. 
Although our ALCNet is around 10 times slower than the purely data-driven FPN due to the proposed local contrast measure layer, it is still among the fastest methods compared with model-driven methods.
However, considering the performance boost brought by the proposed method, we think it is a good trade-off between the performance and inference time. 
In addition, our network can be further accelerated with GPUs. 
\vspace{-.5\baselineskip}

\setlength{\tabcolsep}{4pt}
\begin{table*}[htbp]
\begin{center}
\caption{Comparison with other state-of-the-art methods on IoU and nIoU.}
\label{Tab:IoU}
\vspace{-.5\baselineskip}
\small
\begin{tabular}{Sc Sc Sc Sc Sc Sc Sc Sc Sc Sc Sc Sc} 
\toprule  
Metric        & SMSL  & FKRW  & MPCM  & IPI    & NIPPS & RIPT  & FPN   & SK-FPN & GAU-FPN & TBC-Net & ALCNet  \\
\midrule 
IoU           & 0.081 & 0.268 & 0.357 & 0.466  & 0.473 & 0.146 & 0.720 & 0.702  & 0.701   & 0.734   & \textbf{0.757} \\
nIoU          & 0.279 & 0.339 & 0.445 & 0.607  & 0.602 & 0.245 & 0.700 & 0.695  & 0.702   & 0.713   & \textbf{0.728} \\
Time on CPU/s & 0.595 & 0.399 & 0.347 & 11.699 & 5.707 & 6.398 & 0.031 & 0.035  & 0.033   & 0.049   & 0.378 \\
\bottomrule
\end{tabular}
\end{center}
\vspace*{-2.\baselineskip}
\end{table*}

\subsection{Error Diagnosis and Limitations}

\begin{figure*}[htbp]
  \centering
  \includegraphics[width=.99\textwidth]{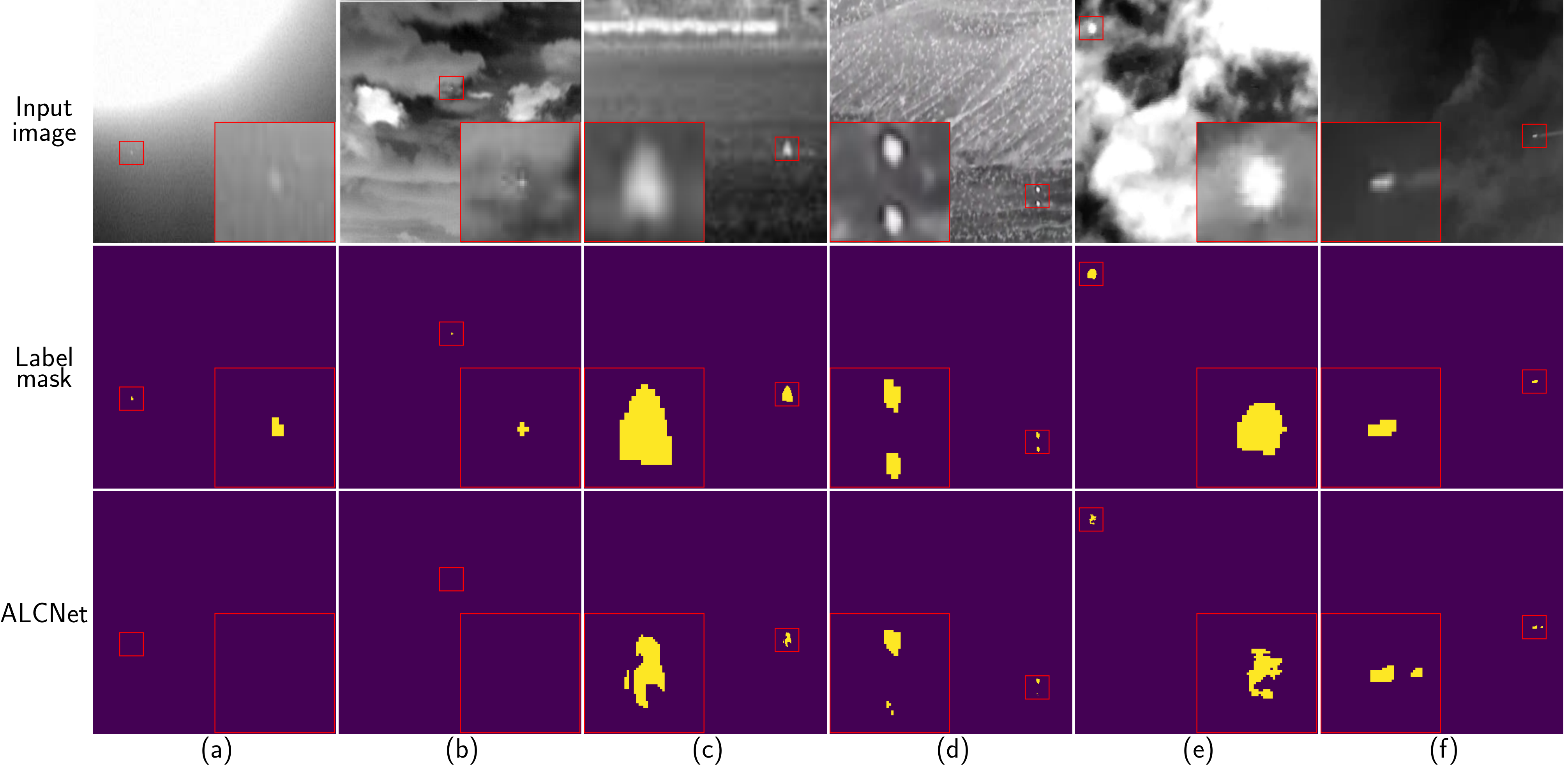}
  \caption{Illustrations of the images on which the proposed ALCNet can not perform very well}
  \label{fig:hard}
  \vspace{-1.\baselineskip}
\end{figure*}

In this part, we analyze the reasons for false positives and false negatives as well as their impact on the detection performance. 
All the prediction results of the test set are available online\footnote{\url{https://github.com/YimianDai/open-alcnet/tree/master/results/pred}}.
The images on which the proposed ALCNet can not perform very well are shown in \cref{fig:hard}. 
Actually, the overall performance of the proposed ALCNet is quite good, only having two miss detections, namely \cref{fig:hard} (a) and (b).
It can be seen that the majority of segmentation errors stems from the target boundary, either exceeding the labeled mask by a few pixels, or segmenting the target incompletely.
It should be noted that the human label is ambiguous or inaccurate in terms of one or two pixels' shift, which has a large impact on our final IoU and nIoU metrics. 
For example, for a minimum target of $2 \times 2$ pixels, even one pixel's shift leads the label to $3 \times 3$, which induces about 50\% errors for this target.
Actually, such boundary error also exists in generic vision tasks. But due to the large object size, the impact is not as prominent as our infrared small target detection task.
However, as long as the segmented pixels are located in the same connected domain, these boundary errors will not cause true false alarms or missed detections.
Further, the false positives caused by incomplete detection like \cref{fig:hard} (c), (d), and (f), in which one target is predicted as two close but separated regions, can be alleviated by simple morphological dilation operations under the target sparsity prior.
The real issue is the miss detections. From \cref{fig:hard} (a) and (b), it can be seen that the main reason is that the infrared small target is too dim. 
Besides, the small size of the target also causes its weight to be very small in the loss function, which is easily overwhelmed by the boundary error of larger targets during training.

It should be noted that in this work, we follow the convention of this field and model infrared small target detection as a segmentation problem. 
Actually, we start our work as a bounding box regression problem just like the conventional object detection task in generic computer vision.
However, due to the small size, the IoU threshold has to be much lower than the default value in generic vision datasets, which leads to a severe duplicate detection problem as another form of the localization error.
Therefore, it remains an open question that how to represent the infrared small target so that the evaluation metric can reveal the true detection performance and not be affected by the boundary error in any means.

\section{Conclusion}

We have presented our ALCNet for infrared small target detection.
In particular, we find that convolutional networks that integrate the local contrast prior, which is generally modeled in non-learning methods, are very promising and worthy of further research.
By breaking the conventional non-overlapped patch constraint, we extract and fuse the local contrast feature maps in two stages, namely in the same layer and across layers, to better transplant the domain knowledge into networks.
Besides, to highlight and preserve the small target in high-level coarse features, we utilize a bottom-up local attentional modulation module that embeds subtle low-level details into high-level layers. 
We conduct extensive ablation studies and comparison with other state-of-the-art methods.
The proposed ALCNet significantly outperforms the compared purely model-driven methods and purely data-driven networks on the open SIRST dataset, which suggests that one should pay attention to combining deep networks with domain knowledge to detect small infrared targets and a target-preserving cross-layer fusion scheme holds the potential to yield better results.

\section*{Acknowledgment}
This work was supported in part by the National Natural Science Foundation of China under Grant No. 61573183, the Open Project Program of the National Laboratory of Pattern Recognition (NLPR) under Grant No. 201900029, the Nanjing University of Aeronautics and Astronautics PhD short-term visiting scholar project under Grant No. 180104DF03, the Excellent Chinese and Foreign Youth Exchange Program, China Association for Science and Technology, China Scholarship Council under Grant No. 201806830039.

\bibliographystyle{IEEEtran}
\bibliography{reference}

\begin{thebibliography}{10}
\providecommand{\url}[1]{#1}
\csname url@samestyle\endcsname
\providecommand{\newblock}{\relax}
\providecommand{\bibinfo}[2]{#2}
\providecommand{\BIBentrySTDinterwordspacing}{\spaceskip=0pt\relax}
\providecommand{\BIBentryALTinterwordstretchfactor}{4}
\providecommand{\BIBentryALTinterwordspacing}{\spaceskip=\fontdimen2\font plus
\BIBentryALTinterwordstretchfactor\fontdimen3\font minus
  \fontdimen4\font\relax}
\providecommand{\BIBforeignlanguage}[2]{{%
\expandafter\ifx\csname l@#1\endcsname\relax
\typeout{** WARNING: IEEEtran.bst: No hyphenation pattern has been}%
\typeout{** loaded for the language `#1'. Using the pattern for}%
\typeout{** the default language instead.}%
\else
\language=\csname l@#1\endcsname
\fi
#2}}
\providecommand{\BIBdecl}{\relax}
\BIBdecl

\bibitem{CVPRW14TIVBenchmark}
Z.~Wu, N.~W. Fuller, D.~H. Theriault, and M.~Betke, ``A thermal infrared video
  benchmark for visual analysis,'' in \emph{2014 {IEEE} Conference on Computer
  Vision and Pattern Recognition ({CVPR}) Workshops, Columbus, OH, USA}, 2014,
  pp. 201--208.

\bibitem{ICNNSP03SPIE}
{Wei Zhang}, {Mingyu Cong}, and {Liping Wang}, ``Algorithms for optical weak
  small targets detection and tracking: Review,'' in \emph{International
  Conference on Neural Networks and Signal Processing, Nanjing, China}, vol.~1,
  2003, pp. 643--647.

\bibitem{TAES88MatchedFiltering}
I.~S. Reed, R.~M. Gagliardi, and L.~B. Stotts, ``{Optical Moving Target
  Detection with 3-D Matched Filtering},'' \emph{IEEE Transactions on Aerospace
  and Electronic Systems}, vol.~24, no.~4, pp. 327--336, Jul 1988.

\bibitem{TIP13IPI}
C.~Gao, D.~Meng, Y.~Yang, Y.~Wang, X.~Zhou, and A.~G. Hauptmann, ``Infrared
  patch-image model for small target detection in a single image,'' \emph{IEEE
  Transactions on Image Processing}, vol.~22, no.~12, pp. 4996--5009, 2013.

\bibitem{TASS87MaxMedian}
G.~{Arce} and M.~{McLoughlin}, ``Theoretical analysis of the max/median
  filter,'' \emph{IEEE Transactions on Acoustics, Speech, and Signal
  Processing}, vol.~35, no.~1, pp. 60--69, 1987.

\bibitem{IJCV04BlobDetection}
K.~Mikolajczyk and C.~Schmid, ``Scale {\&} affine invariant interest point
  detectors,'' \emph{International Journal of Computer Vision}, vol.~60, no.~1,
  pp. 63--86, 2004.

\bibitem{JSTARS18GVF}
D.~{Liu}, L.~{Cao}, Z.~{Li}, T.~{Liu}, and P.~{Che}, ``Infrared small target
  detection based on flux density and direction diversity in gradient vector
  field,'' \emph{IEEE Journal of Selected Topics in Applied Earth Observations
  and Remote Sensing}, vol.~11, no.~7, pp. 2528--2554, 2018.

\bibitem{TGRS13LCM}
C.~L.~P. Chen, H.~Li, Y.~Wei, T.~Xia, and Y.~Y. Tang, ``A local contrast method
  for small infrared target detection,'' \emph{IEEE Transactions on Geoscience
  and Remote Sensing}, vol.~52, no.~1, pp. 574--581, 2014.

\bibitem{TGRS16WLDM}
H.~{Deng}, X.~{Sun}, M.~{Liu}, C.~{Ye}, and X.~{Zhou}, ``Small infrared target
  detection based on weighted local difference measure,'' \emph{IEEE
  Transactions on Geoscience and Remote Sensing}, vol.~54, no.~7, pp.
  4204--4214, 2016.

\bibitem{IPT17NIPPS}
Y.~Dai, Y.~Wu, Y.~Song, and J.~Guo, ``Non-negative infrared patch-image model:
  Robust target-background separation via partial sum minimization of singular
  values,'' \emph{Infrared Physics \& Technology}, vol.~81, pp. 182--194, 2017.

\bibitem{JSTARS17RIPT}
Y.~Dai and Y.~Wu, ``Reweighted infrared patch-tensor model with both nonlocal
  and local priors for single-frame small target detection,'' \emph{IEEE
  Journal of Selected Topics in Applied Earth Observations and Remote Sensing},
  vol.~10, no.~8, pp. 3752--3767, 2017.

\bibitem{IPT14BiTDLMS}
Y.~Zhao, H.~Pan, C.~Du, Y.~Peng, and Y.~Zheng, ``Bilateral two-dimensional
  least mean square filter for infrared small target detection,''
  \emph{Infrared Physics \& Technology}, vol.~65, pp. 17--23, 2014.

\bibitem{TGRS20TopHatReg}
H.~Zhu, S.~Liu, L.~Deng, Y.~Li, and F.~Xiao, ``Infrared small target detection
  via low-rank tensor completion with top-hat regularization,'' \emph{IEEE
  Transactions on Geoscience and Remote Sensing}, vol.~58, no.~2, pp.
  1004--1016, 2020.

\bibitem{TGRS18DECM}
X.~Bai and Y.~Bi, ``{Derivative Entropy-Based Contrast Measure for Infrared
  Small-Target Detection},'' \emph{IEEE Transactions on Geoscience and Remote
  Sensing}, vol.~56, no.~4, pp. 2452--2466, Jan 2018.

\bibitem{CVPR15FCN}
J.~Long, E.~Shelhamer, and T.~Darrell, ``Fully convolutional networks for
  semantic segmentation,'' in \emph{2015 IEEE Conference on Computer Vision and
  Pattern Recognition ({CVPR}), Boston, MA, USA}, 2015, pp. 3431--3440.

\bibitem{MICCAI15UNet}
O.~Ronneberger, P.~Fischer, and T.~Brox, ``U-net: Convolutional networks for
  biomedical image segmentation,'' in \emph{18th International Conference on
  Medical Image Computing and Computer-Assisted Intervention ({MICCAI}),
  Munich, Germany}, 2015, pp. 234--241.

\bibitem{TPAMI13RepresentationLearning}
Y.~Bengio, A.~C. Courville, and P.~Vincent, ``Representation learning: {A}
  review and new perspectives,'' \emph{IEEE Transactions on Pattern Analysis
  and Machine Intelligence}, vol.~35, no.~8, pp. 1798--1828, 2013.

\bibitem{arXiv19TBCNet}
M.~Zhao, L.~Cheng, X.~Yang, P.~Feng, L.~Liu, and N.~Wu, ``Tbc-net: A real-time
  detector for infrared small target detection using semantic constraint,''
  \emph{arXiv preprint arXiv:2001.05852}, 2019.

\bibitem{TGRS17SMSL}
X.~Wang, Z.~Peng, D.~Kong, and Y.~He, ``Infrared dim and small target detection
  based on stable multisubspace learning in heterogeneous scene,'' \emph{IEEE
  Transactions on Geoscience and Remote Sensing}, vol.~55, no.~10, pp.
  5481--5493, 2017.

\bibitem{PR16MPCM}
Y.~Wei, X.~You, and H.~Li, ``Multiscale patch-based contrast measure for small
  infrared target detection,'' \emph{Pattern Recognition}, vol.~58, pp.
  216--226, 2016.

\bibitem{GRSL14ILCM}
J.~Han, Y.~Ma, B.~Zhou, F.~Fan, K.~Liang, and Y.~Fang, ``{A Robust Infrared
  Small Target Detection Algorithm Based on Human Visual System},'' \emph{IEEE
  Geoscience and Remote Sensing Letters}, vol.~11, no.~12, pp. 2168--2172, May
  2014.

\bibitem{CVPR17TinyFaces}
P.~Hu and D.~Ramanan, ``Finding tiny faces,'' in \emph{2017 {IEEE} Conference
  on Computer Vision and Pattern Recognition ({CVPR}), Honolulu, HI, USA},
  2017, pp. 1522--1530.

\bibitem{ICCV17S3FD}
S.~Zhang, X.~Zhu, Z.~Lei, H.~Shi, X.~Wang, and S.~Z. Li, ``{S3FD}: Single shot
  scale-invariant face detector,'' in \emph{2017 {IEEE} International
  Conference on Computer Vision ({ICCV}), Venice, Italy}, Oct 2017.

\bibitem{NIPS18SNIPER}
B.~Singh, M.~Najibi, and L.~S. Davis, ``{SNIPER:} efficient multi-scale
  training,'' in \emph{Annual Conference on Neural Information Processing
  Systems (NeurIPS) 2018, Montr{\'{e}}al, Canada}, 2018, pp. 9333--9343.

\bibitem{CVPR18EncNet}
H.~Zhang, K.~J. Dana, J.~Shi, Z.~Zhang, X.~Wang, A.~Tyagi, and A.~Agrawal,
  ``Context encoding for semantic segmentation,'' in \emph{2018 {IEEE}
  Conference on Computer Vision and Pattern Recognition ({CVPR}), Salt Lake
  City, UT, USA}, 2018, pp. 7151--7160.

\bibitem{ECCV16ContextualPriming}
A.~Shrivastava and A.~Gupta, ``Contextual priming and feedback for faster
  {R-CNN},'' in \emph{14th European Conference on Computer Vision ({ECCV}),
  Amsterdam, The Netherlands}, ser. Lecture Notes in Computer Science, vol.
  9905, 2016, pp. 330--348.

\bibitem{CVPR18SNIP}
B.~Singh and L.~S. Davis, ``An analysis of scale invariance in object detection
  - {SNIP},'' in \emph{2018 IEEE Conference on Computer Vision and Pattern
  Recognition (CVPR), Salt Lake City, UT, USA}, June 2018, pp. 3578--3587.

\bibitem{CVPR17FPN}
T.~Lin, P.~Doll{\'{a}}r, R.~B. Girshick, K.~He, B.~Hariharan, and S.~J.
  Belongie, ``Feature pyramid networks for object detection,'' in \emph{2017
  {IEEE} Conference on Computer Vision and Pattern Recognition ({CVPR}),
  Honolulu, HI, USA}, 2017, pp. 936--944.

\bibitem{BMVC18PAN}
H.~Li, P.~Xiong, J.~An, and L.~Wang, ``Pyramid attention network for semantic
  segmentation,'' in \emph{British Machine Vision Conference ({BMVC}) 2018,
  Newcastle, UK}, 2018, pp. 1--13.

\bibitem{SPL19SkipAttention}
W.~Yuan, S.~Wang, X.~Li, M.~Unoki, and W.~Wang, ``A skip attention mechanism
  for monaural singing voice separation,'' \emph{IEEE Signal Processing
  Letters}, vol.~26, no.~10, pp. 1481--1485, 2019.

\bibitem{CVPR18SENet}
J.~Hu, L.~Shen, and G.~Sun, ``Squeeze-and-excitation networks,'' in \emph{2018
  {IEEE} Conference on Computer Vision and Pattern Recognition ({CVPR}), Salt
  Lake City, UT, USA}, 2018, pp. 7132--7141.

\bibitem{CVPR15GoogLeNet}
C.~Szegedy, W.~Liu, Y.~Jia, P.~Sermanet, S.~E. Reed, D.~Anguelov, D.~Erhan,
  V.~Vanhoucke, and A.~Rabinovich, ``Going deeper with convolutions,'' in
  \emph{2015 IEEE Conference on Computer Vision and Pattern Recognition
  ({CVPR}), Boston, MA, USA}, 2015, pp. 1--9.

\bibitem{NIPS18GENet}
J.~Hu, L.~Shen, S.~Albanie, G.~Sun, and A.~Vedaldi, ``Gather-excite: Exploiting
  feature context in convolutional neural networks,'' in \emph{Annual
  Conference on Neural Information Processing Systems (NeurIPS) 2018,
  Montr{\'{e}}al, Canada}, 2018, pp. 9423--9433.

\bibitem{ICLR16DilatedConv}
F.~Yu and V.~Koltun, ``Multi-scale context aggregation by dilated
  convolutions,'' in \emph{4th International Conference on Learning
  Representations ({ICLR}), San Juan, Puerto Rico}, 2016, pp. 1--10.

\bibitem{NIPS16ERF}
W.~Luo, Y.~Li, R.~Urtasun, and R.~S. Zemel, ``Understanding the effective
  receptive field in deep convolutional neural networks,'' in \emph{Annual
  Conference on Neural Information Processing Systems (NeurIPS) 2016,
  Barcelona, Spain}, 2016, pp. 4898--4906.

\bibitem{ICLR14NiN}
M.~Lin, Q.~Chen, and S.~Yan, ``Network in network,'' in \emph{2nd International
  Conference on Learning Representations ({ICLR}), Banff, AB, Canada}, 2014,
  pp. 1--10.

\bibitem{ICML10ReLU}
V.~Nair and G.~E. Hinton, ``Rectified linear units improve restricted boltzmann
  machines,'' in \emph{the 27th International Conference on Machine Learning
  ({ICML}), Haifa, Israel}, ser. ICML'10, USA, 2010, pp. 807--814.

\bibitem{ICML15BN}
S.~Ioffe and C.~Szegedy, ``Batch normalization: Accelerating deep network
  training by reducing internal covariate shift,'' in \emph{the 32nd
  International Conference on Machine Learning {(ICML)}, Lille, France}, 2015,
  pp. 448--456.

\bibitem{ECCV16ResNetV2}
K.~He, X.~Zhang, S.~Ren, and J.~Sun, ``Identity mappings in deep residual
  networks,'' in \emph{14th European Conference on Computer Vision ({ECCV}),
  Amsterdam, The Netherlands}, 2016, pp. 630--645.

\bibitem{SoftIoU}
M.~A. Rahman and Y.~Wang, ``Optimizing intersection-over-union in deep neural
  networks for image segmentation,'' in \emph{12th International Symposium on
  Visual Computing ({ISVC}), Las Vegas, NV, USA}, 2016, pp. 234--244.

\bibitem{COLT10Adagrad}
J.~C. Duchi, E.~Hazan, and Y.~Singer, ``Adaptive subgradient methods for online
  learning and stochastic optimization,'' in \emph{Conference on Learning
  Theory ({COLT})}, 2010, pp. 257--269.

\bibitem{ICCV15PReLU}
K.~He, X.~Zhang, S.~Ren, and J.~Sun, ``Delving deep into rectifiers: Surpassing
  human-level performance on imagenet classification,'' in \emph{2015 IEEE
  International Conference on Computer Vision ({ICCV}), Santiago, Chile}, ser.
  ICCV '15, Washington, DC, USA, 2015, pp. 1026--1034.

\bibitem{CVPR19SKNet}
X.~Li, W.~Wang, X.~Hu, and J.~Yang, ``Selective kernel networks,'' in
  \emph{2019 {IEEE} Conference on Computer Vision and Pattern Recognition
  ({CVPR}), Long Beach, CA, USA}, 2019, pp. 510--519.

\bibitem{TGRS19FKRW}
Y.~Qin, L.~Bruzzone, C.~Gao, and B.~Li, ``Infrared small target detection based
  on facet kernel and random walker,'' \emph{IEEE Transactions on Geoscience
  and Remote Sensing}, vol.~57, no.~9, pp. 7104--7118, 2019.

\end{thebibliography}

\begin{IEEEbiography}[{\includegraphics[width=1in,height=1.25in,clip,keepaspectratio]{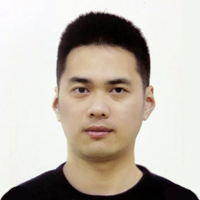}}]{Yimian Dai} 
    received the B.S. degree from Nanjing University of Aeronautics and Astronautics (NUAA), Nanjing, China, in 2013, where he is currently pursuing the Ph.D. degree. 
    As a visiting student, he spent a half year in the department of computer science at the University of Copenhagen and later two years in the department of computer science at the University of Arizona.     
His current interests include sparse representation, deep learning, and their applications in image classification, target detection, and image understanding.
\end{IEEEbiography}

\begin{IEEEbiography}[{\includegraphics[width=1in,height=1.25in,clip,keepaspectratio]{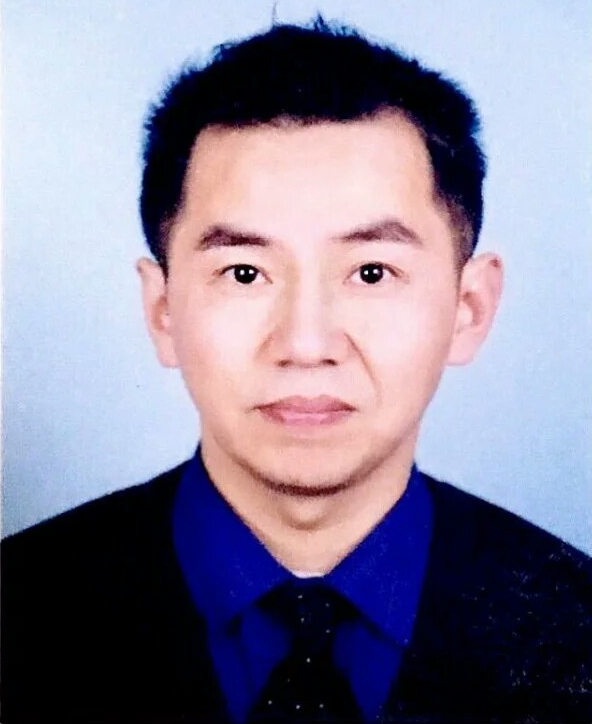}}]{Yiquan Wu}
    received his M.S. and Ph.D. degrees from Nanjing University of Aeronautics and Astronautics in 1987 and 1998, respectively. He is at present a professor and Ph.D. supervisor in the Department of Information and Communication Engineering at the Nanjing University of Aeronautics and Astronautics, where he is involved in teaching and research in the areas of image processing and recognition, target detection and tracking, and intelligent information processing. 
\end{IEEEbiography}

\begin{IEEEbiography}[{\includegraphics[width=1in,height=1.25in,clip,keepaspectratio]{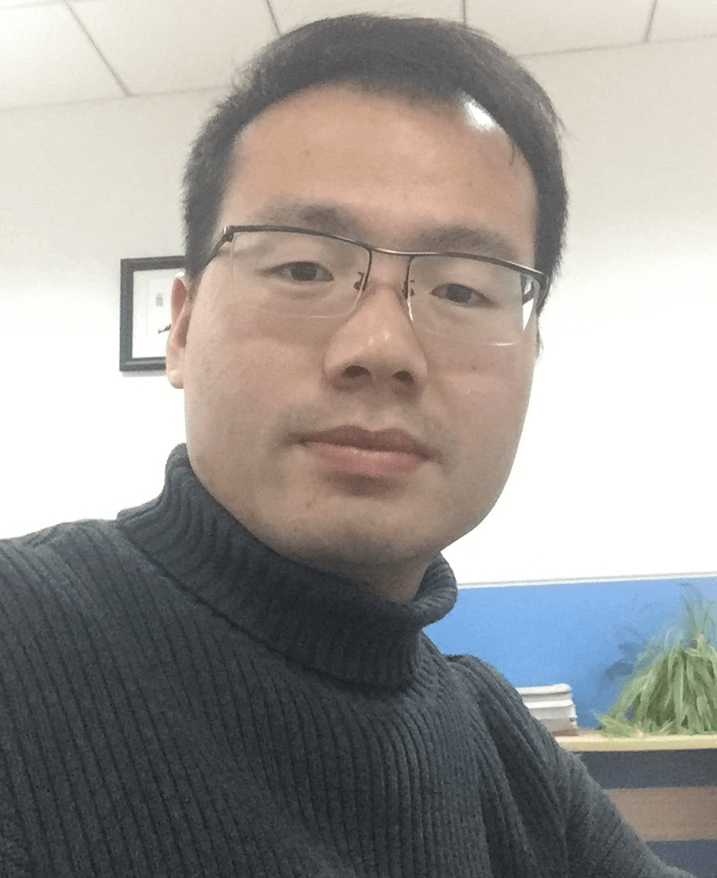}}]{Fei Zhou}
    received his M.S. degree in electronic engineering from Xinjiang University in 2017. Now he is a Ph.D. candidate in College of Electronic and Information Engineering, Nanjing University of Aeronautics and Astronautics. His research interests include signal processing, target detection and image processing.
\end{IEEEbiography}

\begin{IEEEbiography}[{\includegraphics[width=1in,height=1.25in,clip,keepaspectratio]{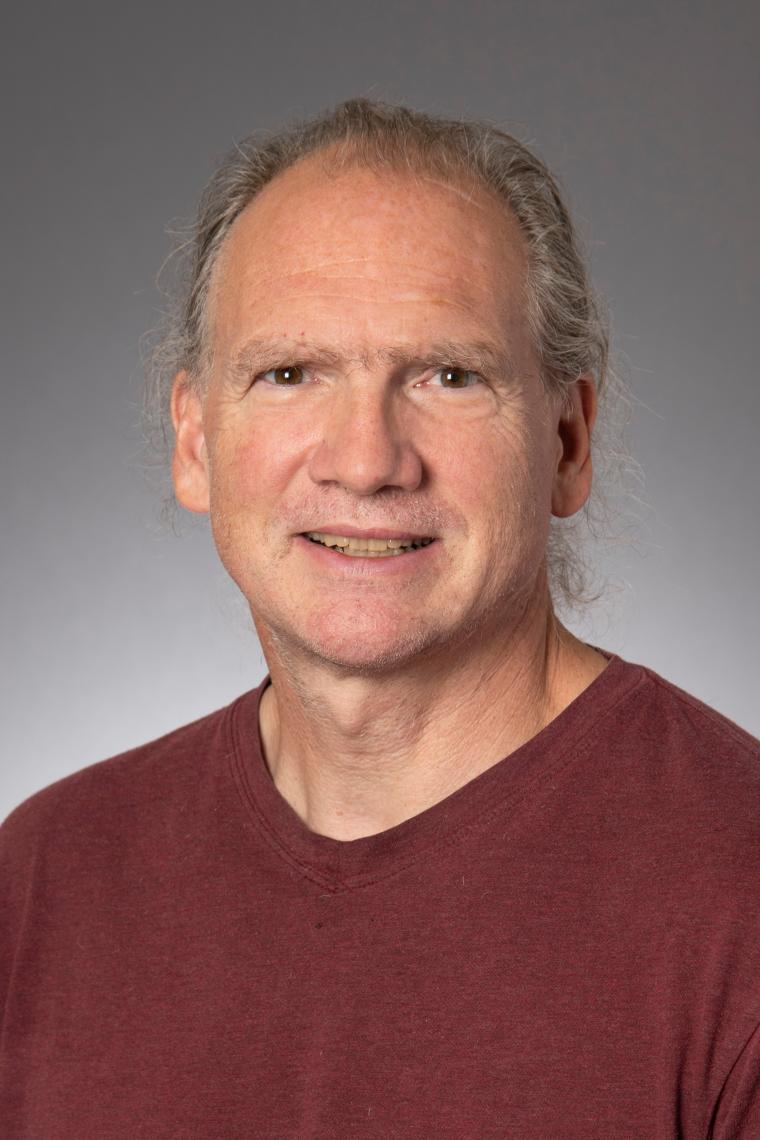}}]{Kobus Barnard} Kobus Barnard received the Ph.D. degree in computer science with specialization
    in computational color constancy from Simon Fraser University, Burnaby, BC,
    Canada, in 2000.  He was awarded the Governor General Gold Medal across all
    disciplines for his Ph.D. dissertation
    
    Dr. Barnard then spent two years at the University of California at Berkeley as
    a Post-Doctoral Researcher, working on modeling the joint statistics of images
    and associated text.
    
    He is currently a Professor with the Department of Computer Science at the
    University of Arizona. He also holds appointments in Statistics, Cognitive
    Science, Electrical and Computer Engineering, and the BIO5 Institute. He leads
    the Interdisciplinary Visual Intelligence Laboratory. His research interests
    include addressing interdisciplinary computational intelligence by developing
    top-down statistical models that are predictive, semantic, and explanatory.
    Application domains include computer vision, multimedia data, biological
    structure and processes, astronomy, and human social interaction. His work has
    been funded by multiple grants from the National Science Foundation, including a
    CAREER Award, and awards from the Defense Advanced Research Projects Agency, the
    Office of Naval Research, the Arizona Biomedical Commission, and the University
    of Arizona BIO5 Institute.
\end{IEEEbiography}

\end{document}